\documentclass[10pt,logo,copyright]{nvidiatechreport}
\usepackage[numbers, sort]{natbib}
\definecolor{citecolor}{HTML}{0071BC}
\definecolor{linkcolor}{HTML}{ED1C24}

\usepackage{graphicx}
\usepackage{amsmath}
\usepackage{amssymb}
\usepackage{booktabs}
\usepackage{tikz}
\usepackage{bbm}
\usepackage{bm}
\usepackage{pifont}
\usepackage{cuted}
\usepackage{colortbl}
\usepackage{algorithm}
\usepackage{algpseudocode}
\usepackage{xifthen}
\usepackage{wrapfig}
\usepackage{comment}
\usepackage{multirow}
\usepackage[percent]{overpic}
\usepackage{adjustbox}
\usepackage{xcolor}
\usepackage{tcolorbox}

\definecolor{Gray}{rgb}{0.92,0.92,0.92}
\definecolor{NvidiaGreen}{RGB}{118, 185, 0}

\usepackage{mathtools, nccmath}

\usepackage[capitalize]{cleveref}
\crefname{section}{Sec.}{Secs.}
\Crefname{section}{Section}{Sections}
\Crefname{table}{Table}{Tables}
\crefname{table}{Tab.}{Tabs.}

\title{MoRight: Motion Control Done Right}

\author{
\textbf{Shaowei Liu}\textsuperscript{1,2*} \quad
\textbf{Xuanchi Ren}\textsuperscript{1} \quad
\textbf{Tianchang Shen}\textsuperscript{1} \quad
\textbf{Huan Ling}\textsuperscript{1} \quad
\textbf{Saurabh Gupta}\textsuperscript{2} \quad
\textbf{Shenlong Wang}\textsuperscript{2} \quad
\textbf{Sanja Fidler}\textsuperscript{1} \quad
\textbf{Jun Gao}\textsuperscript{1} \\

\small \textsuperscript{1}NVIDIA \quad
\textsuperscript{2}University of Illinois Urbana-Champaign \\
\small *Work done during an internship at NVIDIA. \\
{\small
\url{https://research.nvidia.com/labs/sil/projects/moright}}
}

\begin{abstract}
\normalfont\mdseries
Generating motion-controlled videos---where user-specified actions drive physically plausible scene dynamics under freely chosen viewpoints---demands two capabilities: (1) \emph{disentangled motion control}, allowing users to separately control the object motion and adjust camera viewpoint; and (2) \emph{motion causality}, ensuring that user-driven actions trigger coherent reactions from other objects rather than merely displacing pixels. Existing methods fall short on both fronts: they entangle camera and object motion into a single tracking signal and treat motion as kinematic displacement without modeling causal relationships between object motion. We introduce \ours, a unified framework that addresses both limitations through disentangled motion modeling. Object motion is specified in a canonical static-view and transferred to an arbitrary target camera viewpoint via temporal cross-view attention, enabling disentangled camera and object control. We further decompose motion into \emph{active} (user-driven) and \emph{passive} (consequence) components, training the model to learn motion causality from data. At inference, users can either supply active motion and \ours predicts consequences (\emph{forward reasoning}), or specify desired passive outcomes and \ours recovers plausible driving actions (\emph{inverse reasoning}), all while freely adjusting the camera viewpoint. Experiments on three benchmarks demonstrate state-of-the-art performance in generation quality, motion controllability, and interaction awareness.

\vspace{0.5em}
\noindent\textbf{Keywords:} Video Generation; Disentangled Motion Control; Causal Motion Reasoning

\end{abstract}

\begin{document}

\newcommand{\figdir}{src_figs}
\newcommand{\visrow}[2]{%
    \includegraphics[width=0.19\linewidth]{\figdir/#1/#2/step_00.pdf} &
    \includegraphics[width=0.19\linewidth]{\figdir/#1/#2/step_01.pdf} &
    \includegraphics[width=0.19\linewidth]{\figdir/#1/#2/step_02.pdf} &
    \includegraphics[width=0.19\linewidth]{\figdir/#1/#2/step_03.pdf} &
    \includegraphics[width=0.19\linewidth]{\figdir/#1/#2/step_04.pdf}%
}

\newcommand{\tree}{\bm{\Gamma}}

\newcommand{\robotD}[0]{RoboArt\xspace}
\newcommand{\sapiensD}[0]{Sapiens\xspace}
\newcommand{\wim}[0]{WatchItMove\xspace}
\newcommand{\mbs}[0]{MultiBodySync\xspace}
\newcommand{\xpar}[1]{\noindent\textbf{#1}\ \ }
\newcommand{\vpar}[1]{\vspace{3mm}\noindent\textbf{#1}\ \ }

\newcommand{\sect}[1]{Section~\ref{#1}}
\newcommand{\sects}[1]{Sections~\ref{#1}}
\newcommand{\eqn}[1]{Equation~\ref{#1}}
\newcommand{\eqns}[1]{Equations~\ref{#1}}
\newcommand{\fig}[1]{Figure~\ref{#1}}
\newcommand{\figs}[1]{Figures~\ref{#1}}
\newcommand{\tab}[1]{Table~\ref{#1}}
\newcommand{\tabs}[1]{Tables~\ref{#1}}

\newcommand{\ignorethis}[1]{}
\newcommand{\norm}[1]{\lVert#1\rVert}
\newcommand{\fcseven}{$\mbox{fc}_7$}

\renewcommand*{\thefootnote}{\fnsymbol{footnote}}

\def\naive{na\"{\i}ve\xspace}
\def\Naive{Na\"{\i}ve\xspace}

\makeatletter
\DeclareRobustCommand\onedot{\futurelet\@let@token\@onedot}
\def\@onedot{\ifx\@let@token.\else.\null\fi\xspace}

\def\iid{\emph{i.i.d}\onedot}
\def\eg{\emph{e.g}\onedot} \def\Eg{\emph{E.g}\onedot}
\def\ie{\emph{i.e}\onedot} \def\Ie{\emph{I.e}\onedot}
\def\cf{\emph{c.f}\onedot} \def\Cf{\emph{C.f}\onedot}
\def\etc{\emph{etc}\onedot} \def\vs{\emph{vs}\onedot}
\def\wrt{w.r.t\onedot} \def\dof{d.o.f\onedot}
\def\etal{\emph{et al}\onedot}
\makeatother

\definecolor{citecolor}{RGB}{34,139,34}
\definecolor{mydarkblue}{rgb}{0,0.08,1}
\definecolor{mydarkgreen}{rgb}{0.02,0.6,0.02}
\definecolor{mydarkred}{rgb}{0.8,0.02,0.02}
\definecolor{mydarkorange}{rgb}{0.40,0.2,0.02}
\definecolor{mypurple}{RGB}{111,0,255}
\definecolor{myred}{rgb}{1.0,0.0,0.0}
\definecolor{mygold}{rgb}{0.75,0.6,0.12}
\definecolor{myblue}{rgb}{0,0.2,0.8}
\definecolor{mydarkgray}{rgb}{0.66,0.66,0.66}

\newcommand{\myparagraph}[1]{\vspace{-6pt}\paragraph{#1}}

\newcommand{\bbR}{{\mathbb{R}}}
\newcommand{\bK}{\mathbf{K}}
\newcommand{\bX}{\mathbf{X}}
\newcommand{\bY}{\mathbf{Y}}
\newcommand{\bk}{\mathbf{k}}
\newcommand{\bx}{\mathbf{x}}
\newcommand{\by}{\mathbf{y}}
\newcommand{\bhy}{\hat{\mathbf{y}}}
\newcommand{\bty}{\tilde{\mathbf{y}}}
\newcommand{\bG}{\mathbf{G}}
\newcommand{\bI}{\mathbf{I}}
\newcommand{\bg}{\mathbf{g}}
\newcommand{\bS}{\mathbf{S}}
\newcommand{\bs}{\mathbf{s}}
\newcommand{\bM}{\mathbf{M}}
\newcommand{\bw}{\mathbf{w}}
\newcommand{\eye}{\mathbf{I}}
\newcommand{\bU}{\mathbf{U}}
\newcommand{\bV}{\mathbf{V}}
\newcommand{\bW}{\mathbf{W}}
\newcommand{\bn}{\mathbf{n}}
\newcommand{\bv}{\mathbf{v}}
\newcommand{\bq}{\mathbf{q}}
\newcommand{\bR}{\mathbf{R}}
\newcommand{\bi}{\mathbf{i}}
\newcommand{\bj}{\mathbf{j}}
\newcommand{\bp}{\mathbf{p}}
\newcommand{\bt}{\mathbf{t}}
\newcommand{\bJ}{\mathbf{J}}
\newcommand{\bu}{\mathbf{u}}
\newcommand{\bB}{\mathbf{B}}
\newcommand{\bD}{\mathbf{D}}
\newcommand{\bz}{\mathbf{z}}
\newcommand{\bP}{\mathbf{P}}
\newcommand{\bC}{\mathbf{C}}
\newcommand{\bA}{\mathbf{A}}
\newcommand{\bZ}{\mathbf{Z}}
\newcommand{\bff}{\mathbf{f}}
\newcommand{\bF}{\mathbf{F}}
\newcommand{\bo}{\mathbf{o}}
\newcommand{\bc}{\mathbf{c}}
\newcommand{\bT}{\mathbf{T}}
\newcommand{\bQ}{\mathbf{Q}}
\newcommand{\bL}{\mathbf{L}}
\newcommand{\bl}{\mathbf{l}}
\newcommand{\ba}{\mathbf{a}}
\newcommand{\bE}{\mathbf{E}}
\newcommand{\bH}{\mathbf{H}}
\newcommand{\bd}{\mathbf{d}}
\newcommand{\br}{\mathbf{r}}
\newcommand{\bb}{\mathbf{b}}
\newcommand{\bh}{\mathbf{h}}

\newcommand{\btheta}{\bm{\theta}}

\newcommand{\bhh}{\hat{\mathbf{h}}}
\newcommand{\ci}{{\cal I}}
\newcommand{\ct}{{\cal T}}
\newcommand{\co}{{\cal O}}
\newcommand{\ck}{{\cal K}}
\newcommand{\cu}{{\cal U}}
\newcommand{\cS}{{\cal S}}
\newcommand{\cQ}{{\cal Q}}
\newcommand{\cT}{{\cal S}}
\newcommand{\cC}{{\cal C}}
\newcommand{\cE}{{\cal E}}
\newcommand{\cF}{{\cal F}}
\newcommand{\cL}{{\cal L}}
\newcommand{\X}{{\cal{X}}}
\newcommand{\Y}{{\cal Y}}
\newcommand{\cH}{{\cal H}}
\newcommand{\cP}{{\cal P}}
\newcommand{\cN}{{\cal N}}
\newcommand{\cU}{{\cal U}}
\newcommand{\cV}{{\cal V}}
\newcommand{\cX}{{\cal X}}
\newcommand{\cY}{{\cal Y}}
\newcommand{\graph}{{\cal H}}
\newcommand{\bayes}{{\cal B}}
\newcommand{\cx}{{\cal X}}
\newcommand{\cg}{{\cal G}}
\newcommand{\cm}{{\cal M}}
\newcommand{\cM}{{\cal M}}
\newcommand{\cG}{{\cal G}}
\newcommand{\cR}{\cal{R}}
\newcommand{\R}{\cal{R}}
\newcommand{\eig}{\mathrm{eig}}

\newcommand{\D}{{\cal D}}
\newcommand{\bfp}{{\bf p}}
\newcommand{\bfd}{{\bf d}}

\newcommand{\cv}{{\cal V}}
\newcommand{\ce}{{\cal E}}
\newcommand{\cy}{{\cal Y}}
\newcommand{\cz}{{\cal Z}}
\newcommand{\cb}{{\cal B}}
\newcommand{\cq}{{\cal Q}}
\newcommand{\cd}{{\cal D}}
\newcommand{\bcf}{{\cal F}}
\newcommand{\cI}{\mathcal{I}}

\newcommand{\ut}{^{(t)}}
\newcommand{\up}{^{(t-1)}}
\newcommand{\yesmark}{\textcolor{green!70!black}{\ding{51}}}
\newcommand{\nomark}{\textcolor{red!80!black}{\ding{55}}}

\newcommand{\bpi}{\boldsymbol{\pi}}
\newcommand{\bphi}{\boldsymbol{\phi}}
\newcommand{\bPhi}{\boldsymbol{\Phi}}
\newcommand{\bmu}{\boldsymbol{\mu}}
\newcommand{\bSigma}{\boldsymbol{\Sigma}}
\newcommand{\bGamma}{\boldsymbol{\Gamma}}
\newcommand{\bbeta}{\boldsymbol{\beta}}
\newcommand{\bomega}{\boldsymbol{\omega}}
\newcommand{\blambda}{\boldsymbol{\lambda}}
\newcommand{\bkappa}{\boldsymbol{\kappa}}
\newcommand{\btau}{\boldsymbol{\tau}}
\newcommand{\balpha}{\boldsymbol{\alpha}}
\def\bgamma{\boldsymbol\gamma}

\newcommand{\prox}{{\mathrm{prox}}}

\newcommand{\pardev}[2]{\frac{\partial #1}{\partial #2}}
\newcommand{\dev}[2]{\frac{d #1}{d #2}}
\newcommand{\dw}{\delta\bw}
\newcommand{\lab}{\mathcal{L}}
\newcommand{\unlab}{\mathcal{U}}
\newcommand{\ind}{1{\hskip -2.5 pt}\hbox{I}}
\newcommand{\ff}[2]{   \cf_{\prec (#1 \rightarrow #2)}}
\newcommand{\vv}[2]{   \cv_{\prec (#1 \rightarrow #2)}}
\newcommand{\dd}[2]{   \delta_{#1 \rightarrow #2}}
\newcommand{\ld}[2]{   \lambda_{#1 \rightarrow #2}}
\newcommand{\en}[2]{  \bD(#1|| #2)}
\newcommand{\ex}[3]{  \bE_{#1 \sim #2}\left[ #3\right]}
\newcommand{\exd}[2]{  \bE_{#1 }\left[ #2\right]}

\newcommand{\se}[1]{\mathfrak{se}(#1)}
\newcommand{\SE}[1]{\mathbb{SE}(#1)}
\newcommand{\SO}[1]{\mathbb{SO}(#1)}

\newcommand{\poselow}{\xi}
\newcommand{\pose}{\bm{\poselow}}
\newcommand{\linpose}{\pose^\ell}
\newcommand{\cbpose}{\pose^c}
\newcommand{\rateparam}{v_i}
\newcommand{\bapose}{\bm{\poselow}_i}
\newcommand{\trackingpose}{\bm{\poselow}}
\newcommand{\rotlow}{\omega}
\newcommand{\rot}{\bm{\rotlow}}
\newcommand{\translow}{v}
\newcommand{\trans}{\bm{\translow}}
\newcommand{\hnorm}[1]{\left\lVert#1\right\rVert_{\gamma}}
\newcommand{\lnorm}[1]{\left\lVert#1\right\rVert}
\newcommand{\barate}{v_i}
\newcommand{\trackingrate}{v}
\newcommand{\imgpt}{\mathbf{u}_{i,k,j}}
\newcommand{\mappt}{\mathbf{X}_{j}}
\newcommand{\timet}[1]{\bar{t}_{#1}}
\newcommand{\mf}[1]{\text{MF}_{#1}}
\newcommand{\kmf}[1]{\text{KMF}_{#1}}
\newcommand{\Exp}{\text{Exp}}
\newcommand{\Log}{\text{Log}}

\newcommand{\ours}{MoRight\xspace}

\maketitle

\begin{center}
\centering
\vspace{-1em}
\includegraphics[width=\linewidth]{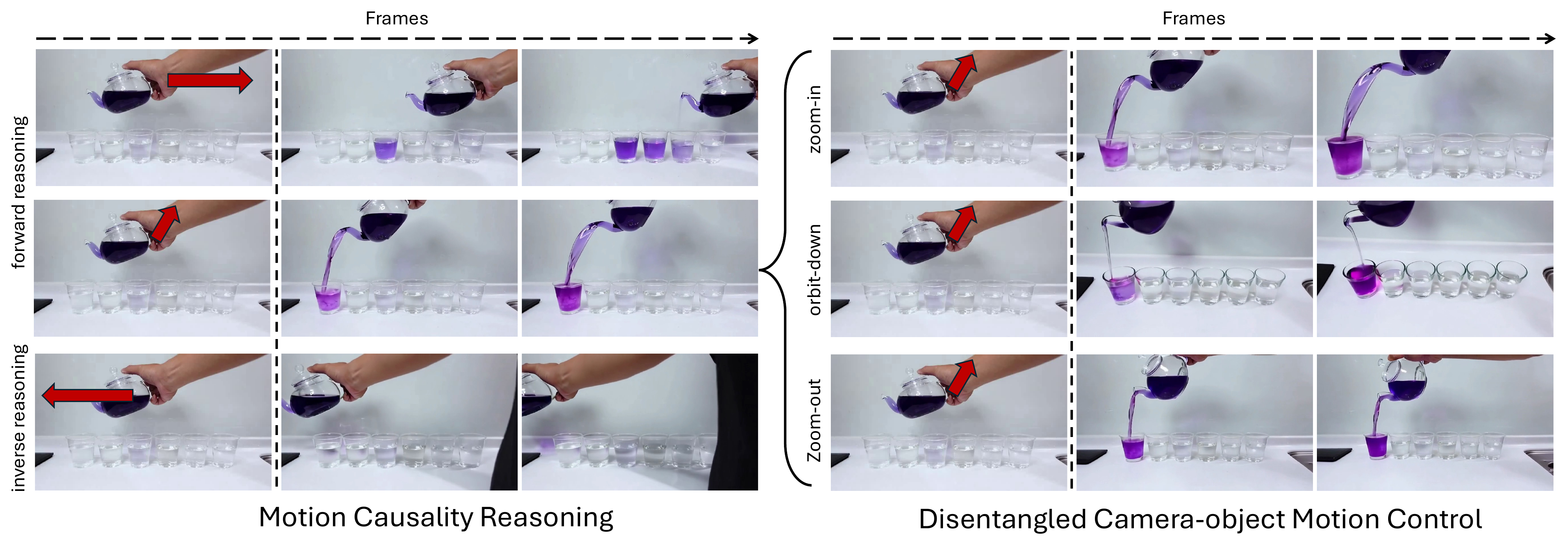}
\vspace{-1.5em}
\captionof{figure}{Given a single input image, our method enables controllable interactive motion generation with motion causality reasoning.
\textit{Left:} Users can provide active motion (\eg action of hand) to drive scene dynamics (\emph{forward reasoning}) or specify desired passive outcomes (\eg trajectory of teapot) and recover plausible driving actions (\emph{inverse reasoning}).
\textit{Right:} The model further enables disentangled control of object motion and camera viewpoint, allowing users to explore the scene with custom viewpoints and motions.}
\label{fig:teaser}
\end{center}

\abscontent

\section{Introduction}
\label{sec:intro}

Humans interact with the physical world as active agents: we move our viewpoint, manipulate objects, and reason about how actions lead to consequences. Yet existing video generation models lack this unified capability~\cite{wan2025wan, CogVideo, AlignYourLatents, CogVideoX, VideoCrafter2}. Bridging this gap is essential for applications that demand interactive visual reasoning, from embodied AI agents~\cite{bar2025navigation, yang2023learning, wang2024drivedreamer} that must anticipate action outcomes~\cite{hafner2019dream, hafner2023mastering, gao2026dreamdojo}, to world models~\cite{hong2025relic, Sora, he2025matrix, yang2025matrix, ye2026world} that simulate physical interactions, and to immersive content creation~\cite{liu2025ponimator, liu2024physgen, chen2025physgen3d, li2025wonderplay, gao2024gaussianflow, wu2024draganything, zhang2024physdreamer} where users freely navigate and manipulate scenes. A desirable video generation system should therefore offer joint controllability over both camera and object motion---generating visually coherent frames under arbitrary viewpoint changes while producing causally consistent scene dynamics driven by user-specified actions.

Existing controllable video generation methods~\cite{MotionCtrl, geng2024motionprompting, wang2025ati, chu2025wanmove, burgert2025go, wang2024boximator} work as {\it renderers}: given displacements for {\it all} pixels, they generate a visually realistic video that adheres to the displacements.
These approaches have two key limitations in practice. First, they entangle camera and object motion, making joint control ambiguous because viewpoint changes alter pixel trajectories. Second, they interpret user-specified motion as simple kinematic displacement and largely ignore the consequences of the given trajectories. Models, therefore, focus on following trajectories rather than reasoning about causal relationships between objects. In reality, actions cause consequences—pushing a cup may cause it to slide and collide with other objects, while lifting a teapot may cause water to pour. Specifying the effects of all the objects' motion through motion prompts is often impractical. Without modeling these action–consequence relationships, motion-controlled generation cannot capture the causal structure of real-world interactions.

To overcome these challenges, we introduce \ours, a unified framework for video generation with disentangled camera–object motion control and motion causality reasoning as shown in \cref{fig:teaser}.
Given a reference image, user-specified motion trajectories, and target viewpoints, \ours generates videos where objects follow the desired motion and the scene is rendered from the specified cameras.
For disentangled camera-object motion control, our key insight is that specifying the motion of objects under camera changes is inherently difficult. We therefore introduce a \textbf{dual-stream motion formulation}. The first branch models and generates object motion in the source image plane under a canonical static viewpoint, allowing users to easily specify dynamic trajectories. The second branch represents the target camera motion and transfers object dynamics from the canonical branch via temporal cross-view attention. This \emph{cross-view motion transfer} enables independent control of camera and object motion while maintaining coherent scene dynamics.
For motion causality reasoning,
we achieve this by decomposing object motion into two categories during training: \emph{active motion}, representing user-driven actions, and \emph{passive motion}, representing their causal outcomes. By conditioning on either active or passive motion signals, the video model learns to generate all the scene dynamics, capturing action--consequence relationships within the scene. This yields two complementary capabilities: forward reasoning of scene evolution from user actions, and inverse reasoning of plausible actions that produce a desired outcome.

We evaluate \ours on three benchmarks covering diverse interaction scenarios. Results show that \ours outperforms existing methods in generation quality, motion controllability, and interaction awareness, validating the effectiveness of disentangled motion control and causal motion reasoning.

In summary, our contributions are threefold.
(1)~We propose a disentangled framework for camera and object motion control, enabling users to draw motion trajectories directly in the image plane while freely adjusting viewpoints to generate coherent videos.
(2)~We empower video generation models with motion reasoning capability by modeling action--consequence relationships, allowing user-driven motions to meaningfully interact with the environment and produce consistent scene dynamics.
(3)~We demonstrate that \ours supports both forward and inverse reasoning: given active motion inputs, it predicts future scene evolution; given desired passive outcomes, it recovers plausible actions that achieve them. Together, these advances establish a unified framework for controllable and reasoning-aware video generation.

\section{Related Work}
\label{sec:related}

\subsection{Motion-Controlled Video Generation}
Controllable video generation has been studied across a spectrum of motion granularity, from coarse region-level signals such as bounding boxes~\cite{wang2024boximator, ma2023trailblazer, xing2025motioncanvas}, sparse keypoint tracks~\cite{yin2023dragnuwa, MotionCtrl,wu2024draganything, li2025magicmotion}, optical flow fields~\cite{jin2025flovd, NSFF, zhang2025motionpro, FOMM, shi2024motion}, and dense per-pixel trajectories~\cite{geng2024motionprompting,chu2025wanmove}. Despite steady progress, trajectory-based methods share two fundamental limitations. First, because trajectories are defined in pixel space, they inevitably entangle object and camera motion: any viewpoint change alters all trajectories, making joint control ill-posed without explicit foreground--background decomposition. Second, producing physically plausible motion typically demands carefully crafted trajectories from dedicated motion-generation models~\cite{liu2025ponimator, lv2024gpt4motion, liu2024physgen, shi2024motion, chen2025physgen3d, montanaro2024motioncraft} or laborious manual annotation. \ours overcomes both issues by disentangling camera and object motion by design, and by accepting lightweight inputs---simple strokes or sparse tracklets---that the model completes into coherent, interaction-aware dynamics.

\subsection{Camera--Object Motion Disentanglement}
Separating camera motion from scene dynamics~\cite{huang2025vipe, li2025megasam, zhang2022structure, kopf2021robust, yao2025uni4d} is a long-standing challenge in controllable generation~\cite{MoCoGAN, Inmodegan, G3AN, geng2024motionprompting} and video understanding~\cite{liu2025visual, huang2025segment, wumotion}. Existing methods~\cite{zhang2025motionpro, gu2025diffusion, chen2025perception, shi2024motion} attempt to decouple the two by treating them as independent control signals, yet they typically rely on privileged information such as per-frame depth~\cite{gu2025diffusion, chen2025perception}, 3D object trajectories~\cite{karaev23cotracker, doersch2023tapir, harley2025alltracker}, or foreground--background segmentation masks~\cite{zhang2025motionpro, liang2024flowvid, niu2024mofa, shi2024motion}, and pre-warp all signals to their anticipated future locations. These assumptions implicitly require the full video sequence or 3D motion to be known in advance, severely limiting applicability in image-to-video settings where only a single reference frame is available. \ours instead introduces a canonical static-view branch for object dynamics and transfers them to arbitrary target viewpoints via cross-view attention, eliminating the need for explicit 3D supervision or pre-computed scene decomposition.

\subsection{Interaction and Causal Reasoning in Video Generation}
A complementary line of work seeks to move beyond kinematic animation toward causally grounded video generation. One family of methods incorporates external physics engines~\cite{liu2024physgen, li2025wonderplay, chen2025physgen3d, montanaro2024motioncraft} or conditions on explicit action representations such as force vectors~\cite{gillman2025force} to model specific phenomena (e.g., fluid flow, rigid-body collisions). While effective in constrained domains, these approaches are tailored to particular interaction types and require a simulation module in the loop, limiting their generality. Another family delegates causal reasoning to vision--language or large language models~\cite{yang2025vlipp, pan2024vlp, lian2023llm, wu2024self}, which first predict outcomes in text and then transfer them to the video generator through intermediate representations such as flow fields~\cite{lv2024gpt4motion, montanaro2024motioncraft}, edge maps~\cite{lv2024gpt4motion}, or depth~\cite{lv2024gpt4motion, zhang2023adding}. This two-stage pipeline is prone to error propagation: imprecise textual predictions are further degraded during cross-modal conversion, yielding spatially inaccurate dynamics. \ours sidesteps both limitations by learning latent action--response structure directly from video data. Decomposing motion into active (user-driven) and passive (consequence) components allows the model to perform cause--effect reasoning at the pixel level within a single forward pass, enabling both forward prediction of scene consequences and inverse inference of plausible underlying actions.

\section{Approach}
\label{sec:approach}

\begin{figure}
\centering
\includegraphics[width=\linewidth, trim=0 0 1em 0, clip]{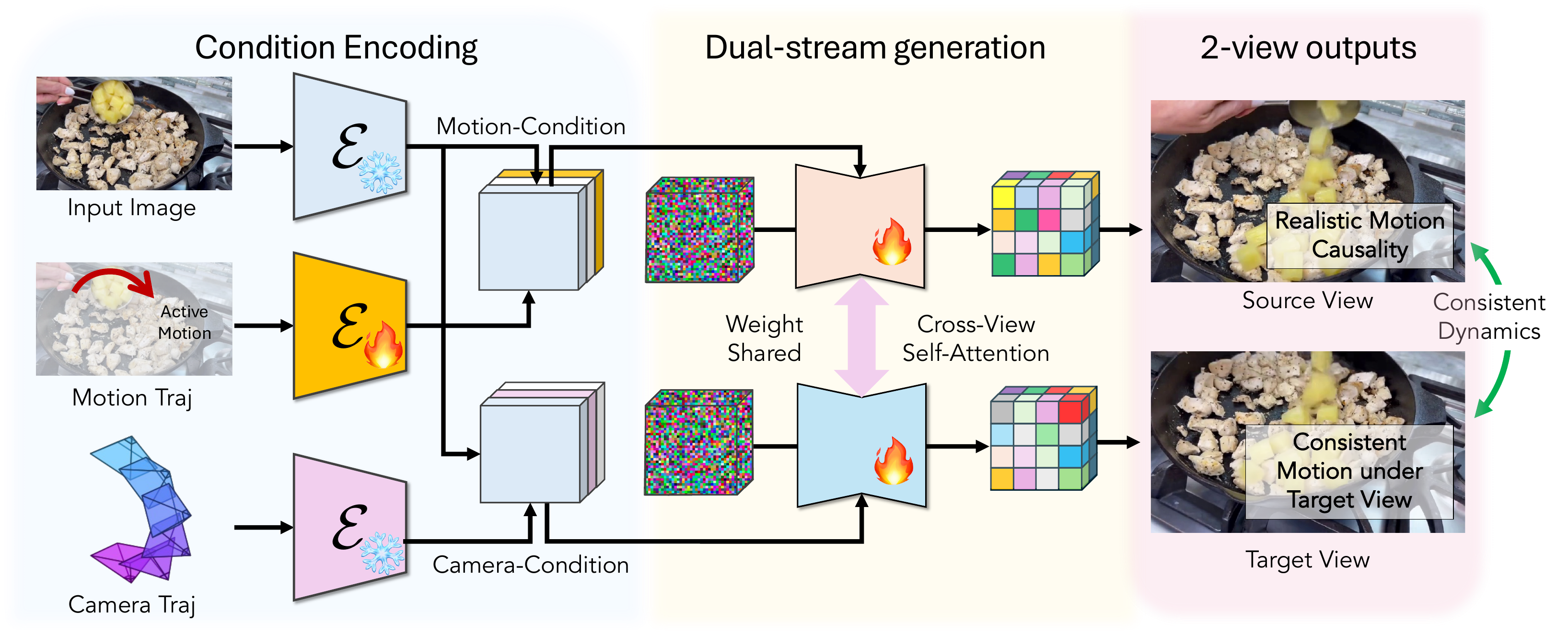}
\caption{\textbf{Model architecture.} Our model adopts a dual-stream architecture with shared weights to disentangle object motion from camera motion. The canonical stream encodes motion trajectories using a track encoder and learns motion in a fixed canonical view. The target stream encodes camera pose signals through a camera encoder. The resulting motion and camera conditions are injected into every attention block of the network. Cross-view self-attention connects the two streams, transferring motion learned in the canonical view to the target view and enabling disentangled camera–object motion generation.}
\label{fig:model}
\end{figure}

Given a single image $I$, we aim to generate a video of $T$ frames $\mathbf{x} \in \mathbb{R}^{T \times H \times W \times 3}$ that follows the user-defined motion of an object represented as pixel trajectories $\mathcal{T}$, and the specified camera motion sequence $\{C_i\}_{i=1}^{T}$. The generated video should be able to model the causality of motion and produce a coherent dynamics within the scene.
To achieve this, we first present our approach for disentangled camera and object motion control in~\cref{sec:disentangle} and \cref{fig:model}, and then introduce motion causality modeling in~\cref{sec:interaction}.
\cref{sec:data} describes our training data curation pipeline, and training details along with the inference pipeline are provided in~\cref{sec:training}.

\subsection{Disentangled Camera-Object Motion Control}
\label{sec:disentangle}
Most existing approaches adopt pixel-wise trajectories as motion control signals. However, such representations inherently entangle object motion with viewpoint changes. The video model must implicitly reason how an object moves, and the camera changes, without explicit geometric cues, when conditioned on these signals.
Our key insight is that object motion is intrinsically unambiguous when expressed in a canonical camera. Building on this observation, we decouple the motion signal from the viewpoint transformation through a dual-stream generation framework~\cite{bai2025recammaster, bai2024syncammaster}. The first stream synthesizes a canonical video in a static camera, where object motion can be directly and faithfully controlled. The second stream generates the target video with both camera and object motion. The two streams can interact with each other through self-attention as shown in \cref{fig:model}.
By jointly denoising both streams, the model learns to transfer motion cues from canonical space to arbitrary camera poses, where the canonical stream serves as an anchor for motion control. This design resolves motion–camera entanglement at generation time while naturally supporting heterogeneous supervision—including motion-only, camera-only, and fully coupled data—as detailed in~\cref{sec:data}.

\noindent\textbf{Preliminaries.}\quad
We build upon a DiT-based~\cite{peebles2023scalable} latent video diffusion models~\cite{agarwal2025cosmos,wan2025wan}.
A pretrained VAE encoder first encodes the video into a latent space
$\mathbf{z}_0 = \mathcal{E}(\mathbf{x}) \in \mathbb{R}^{\hat{T} \times
\hat{H} \times \hat{W} \times d}$.
The diffusion model is trained in this latent space via flow matching~\cite{lipman2022flow}. Specifically, we first sample a noise from a Gaussian distribution: $\boldsymbol{\epsilon} \sim \mathcal{N}(0,\mathbf{I})$, and form $\mathbf{z}_t = (1{-}t)\,\mathbf{z}_0 + t\,\boldsymbol{\epsilon}$ for $t \in [0,1]$. The DiT $\mathcal{G}_\theta$  is trained to regress the velocity:
\begin{equation}
    \mathcal{L} = \mathbb{E}_{\mathbf{z}_0,\, t,\, \boldsymbol{\epsilon}} \left[\left\| \mathcal{G}_\theta(\mathbf{z}_t, t, \mathbf{c}) - (\boldsymbol{\epsilon} - \mathbf{z}_0) \right\|^2 \right],
    \label{eq:flow_matching}
\end{equation}
where $\mathbf{c}$ denotes conditioning signals (\eg, text).
At inference, an ODE solver~\cite{zhao2023unipc} integrates the learned
velocity from a noise to a clean latent $\hat{\mathbf{z}_0}$,  from which a decoder reconstructs the original video $\hat{\mathbf{x}} = \mathcal{D}(\hat{\mathbf{z}_0})$.

\noindent\textbf{Dual-stream generation.}\quad
The user first provides the object motion $\{\boldsymbol{\tau}^{\text{can}}_i\}_{i=1}^T$ in the canonical frame. The first stream generates the videos only with object motion, and the second stream generates the videos with both object and camera motion.
Concretely, the two streams receive their individual conditions:
\begin{equation*}
  \mathbf{c}^{\text{can}} = \bigl\{
    I,\; C_1,\; \{\boldsymbol{\tau}^{\text{can}}_i\}
  \bigr\},
  \qquad
  \mathbf{c}^{\text{tar}} = \bigl\{
    I,\; \{C_i\},\; \emptyset
  \bigr\},
\end{equation*}
where $C_1$ denotes the camera in the first image and is an identity matrix,  and $\emptyset$ denotes empty object motion.

In the following, we assume we have the paired training data: one ground truth video $\mathbf{x}^{\text{can}}$  with object motion only, and the corresponding video $\mathbf{x}^{\text{tar}}$ with both object and camera motion. Extensions to other training data are described in~\cref{sec:data} and~\cref{sec:training}.
We add independent noise with the same timestep $t$ to each stream and obtain $\mathbf{z}^{\text{can}}_t$ and $\mathbf{z}^{\text{tar}}_t$.
We then concatenate two latents along the temporal
dimension and jointly denoise them.
We slightly modify the positional embedding to indicate the difference between the two streams, with details in the supplement.
In this way, we reuse the same DiT weights for two streams, and the only difference is their input and conditioning. The two streams naturally exchange information in the self-attention layers of each transformer block.
During inference, the two streams are jointly denoised, and we provide the output from the target stream  $\mathbf{x}=
\mathcal{D}(\hat{\mathbf{z}}^{\text{tar}}_0)$ to the user, and the canonical stream serves as a ``virtual'' anchor.

\noindent\textbf{Condition injection and motion transfer.}\quad
We inject camera and motion conditions into the latents of DiT at every transformer block. Specifically:

\emph{Camera encoding.} We follow Gen3C~\cite{zhang2023scenewiz3d} and warp the first image $I$  using the corresponding camera pose and estimated depth~\cite{lin2025depth}. We then encode the warped frames via encoder $\mathcal{E}$ from VAE and obtain the latent $\mathbf{z}^{\text{cam}} \in \mathbb{R}^{\hat{T}\times \hat{H}\times\hat{W} \times d}$. For the canonical stream, we use the identity matrix for warping.

\emph{Motion encoding.} Following~\cite{geng2024motionprompting}, we build a per-pixel
trajectory map where pixels along one trajectory share the same temporal-correspondence
embedding. We then encode it via a lightweight encoder to obtain
$\mathbf{e}^{\text{trk}} \in \mathbb{R}^{\hat{T} \times
\hat{H}\times \hat{W} \times d}$. For the target stream, we simply set
$\mathbf{e}^{\text{trk}} = \mathbf{0}$ since the condition is empty.

\emph{Condition injection.} The camera and motion encodings are fused via learned linear projections and added into the latent feature at each transformer block. Let $\mathbf{f}$ denote the feature of one block:
\begin{equation}
  \mathbf{f}^{i} \leftarrow
    \mathbf{f}^{i}
    + W_{\text{cam}}\,\mathbf{z}^{i, \text{cam}}
    + W_{\text{trk}}\,\mathbf{e}^{i, \text{trk}},
  \quad i \in \{\text{can},\, \text{tar}\}.
  \label{eq:cond_inject}
\end{equation}
The features from the two streams are then concatenated  and passed through the self-attention layer:
\begin{equation}
  \bigl[\,\mathbf{f}^{\text{can}};\;
    \mathbf{f}^{\text{tar}}\,\bigr]
  := \operatorname{SelfAttn}\!\bigl(
    \bigl[\,\mathbf{f}^{\text{can}};\;
      \mathbf{f}^{\text{tar}}\,\bigr]\bigr),
  \label{eq:motion_sync}
\end{equation}
allowing target-view tokens to attend to motion-conditioned canonical tokens and vice versa, implicitly exchanging the motion information in latent space. This inject-then-synchronize repeats at every block, progressively transferring motion across views, as shown in \cref{fig:model}.

\subsection{Motion Causality Modeling}
\label{sec:interaction}

\begin{wrapfigure}{r}{0.4\linewidth}
\centering
\vspace{-5em}
\includegraphics[width=\linewidth]{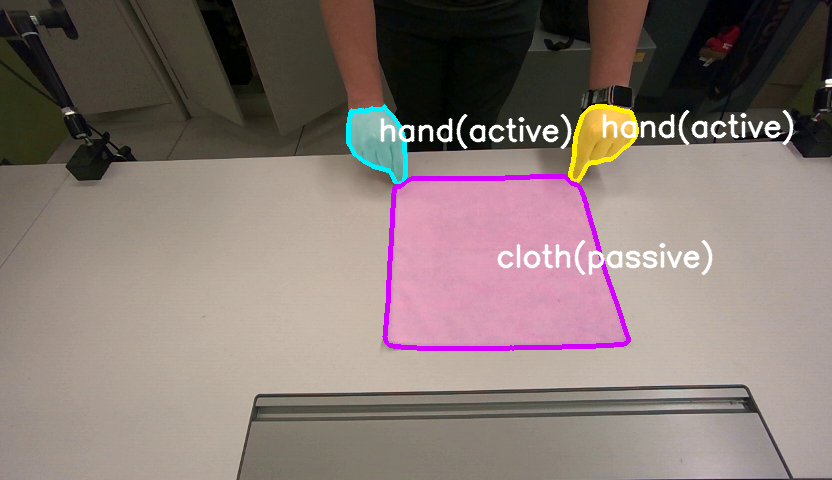}
\caption{Active vs.\ passive motion. The \emph{active} object (hand) initiates the action, while the \emph{passive} object (cloth) responds.}
\label{fig:example}
\vspace{-2em}
\end{wrapfigure}

Disentangling the camera from motion alone is insufficient for realistic interactions: when a hand pushes a cup, the cup must slide; when a ball
strikes a stack of blocks, the blocks must scatter. We term this as
\emph{motion causality}: the ability to reason plausible consequences
 from the given actions, and vice versa.

To model this, we decompose the motion tracks of all foreground objects
$\mathcal{T}=\{\boldsymbol{\tau}_i\}_{i=1}^{T}$ into two complementary components as shown in~\cref{fig:example}:
$\boldsymbol{\tau}_i
  = \boldsymbol{\tau}^{\text{act}}_i \cup \boldsymbol{\tau}^{\text{pas}}_i,
$
where $\boldsymbol{\tau}^{\text{act}}_i$ captures the \emph{active} (causal)
motion---the intentional action applied to the scene (\eg, a hand
pushing)---and $\boldsymbol{\tau}^{\text{pas}}_i$ captures the \emph{passive}
(consequential) motion---the reaction from other objects (\eg, the pushed object sliding). Reasoning such causality is critical in applications such as embodied AI~\cite{bu2025agibot, hafner2019learning, finn2017deep}.

The central mechanism to enable the causality modeling is through \emph{motion dropout}. Specifically,  during training, we
randomly drop out one motion component from the input, and  supervise the model on the full video containing both active and passive motion:
\begin{equation}
  \tilde{\boldsymbol{\tau}}_i :=
  \begin{cases}
    \boldsymbol{\tau}^{\text{act}}_i, & \xi < p, \\[4pt]
    \boldsymbol{\tau}^{\text{pas}}_i, & \text{otherwise},
  \end{cases}
  \label{eq:motion_dropout}
\end{equation}
where $\xi$ is sampled from a uniform distribution $\mathcal{U}(0,1)$, $p$ is the dropout probability and $\tilde{\boldsymbol{\tau}}_i$ is used as tracking condition to video model.

In our model training, we do not distinguish active motion or passive motion when feeding them into the model, and only rely on the model's capability to reason the
dropped component to generate plausible videos.
During training, this asymmetric supervision encourages the video models to internalize the causal relationship between actions and their consequences, rather than simply replaying the provided trajectories.

At inference, this learned causality enables two complementary applications:
\emph{forward reasoning} (action $\to$ reaction), where users specify an
action and the model generates the resulting consequences; and
\emph{inverse reasoning} (reaction $\to$ action), where users prescribe a desired outcome and the model synthesizes a plausible action that drives it. We demonstrate these capabilities in~\cref{sec:exp_casual}.

\begin{figure}
\centering
\includegraphics[width=\linewidth]{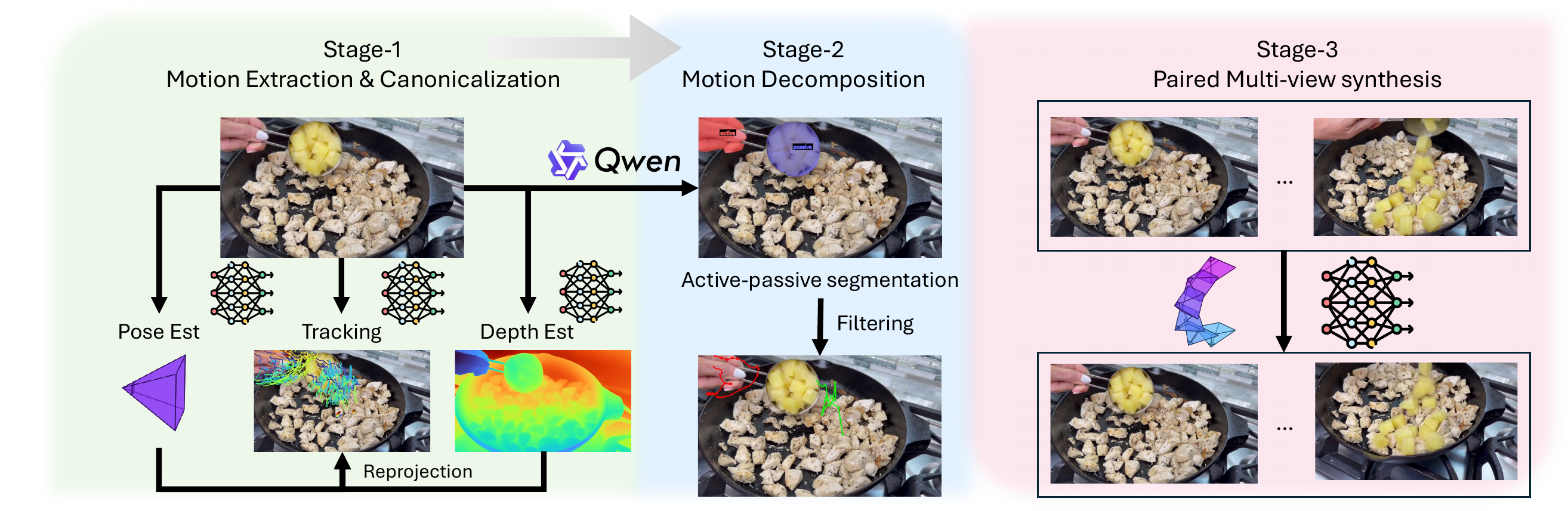}
\caption{\textbf{Data curation pipeline.} Foundation models~\cite{harley2025alltracker, huang2025vipe, ravi2024sam2} extract depth, camera poses, and tracks from raw videos. A VLM~\cite{Qwen3-VL} segments tracks into active/passive regions. We further optionally use a video-to-video model~\cite{fu2026plenoptic} to generate paired videos with the same object motion but different camera motions.}
\label{fig:data_pipeline}
\end{figure}

\subsection{Training Data Curation}
\label{sec:data}
Curating data to train our dual-stream model is challenging since most real-world videos are single-view and always entangle camera and object motion, while our model requires paired videos depicting the same dynamics under different viewpoints. We first describe our data annotation pipeline to extract the pixel trajectories,  camera poses, and active/passive motion, and provide details of our data curation for training afterwards. The overview of pipeline is shown in \cref{fig:data_pipeline}.

\noindent\textbf{Motion extraction and canonicalization.}\quad
Given a video $\mathbf{x}$, we estimate per-frame depth maps $\{D_i\}$, camera poses $\{C_i\}$, and intrinsics~$K$ using ViPE~\cite{huang2025vipe}, and extract dense pixel trajectories $\mathcal{T}=\{\boldsymbol{\tau}_i\}_{i=1}^{T}$ with AllTracker~\cite{harley2025alltracker}.
Each trajectory is unprojected to 3D and reprojected into the first frame:
\begin{equation}
  \boldsymbol{\tau}^{\text{can}}_i
  = \pi\!\bigl(
    K,\, C_0\, C_u^{-1}\,
    \pi^{-1}(K,\, \boldsymbol{\tau}_i,\, D_i)
  \bigr),
  \label{eq:reproject}
\end{equation}
where $\pi^{-1}$ lifts 2D points to 3D using depth and intrinsics, and
$\pi$ projects onto the image plane of~$I_0$. We assume constant intrinsics across the video.

\noindent\textbf{Active and passive motion decomposition.}\quad
For the given video, we prompt a vision-language model (Qwen3~\cite{Qwen3-VL}) to identify the active and passive objects, then segment the first frame with SAM2~\cite{ravi2024sam2}, yielding masks $M^{\text{act}}$ and $M^{\text{pas}}$, for active and passive objects, respectively.
Trajectories are assigned to each component by mask membership, producing $\boldsymbol{\tau}^{\text{can,act}}$ and $\boldsymbol{\tau}^{\text{can,pas}}$
for the motion dropout training in~\cref{eq:motion_dropout}.
We also generate per-video captions describing each motion component and only provide the caption of one component during training to prevent information leakage.

\noindent\textbf{Synthetic data generation for paired two-view videos.}\quad
We leverage a synthetic data generation pipeline to generate paired two-view videos for training. Specifically, we first curate the videos whose cameras are static by checking the displacement of camera poses estimated from ViPE~\cite{huang2025vipe}.
We then synthesize corresponding moving-camera videos using a camera-control video-to-video model~\cite{fu2025plenoptic}, providing supervision for the second stream.
To increase camera diversity, we further augment the data with basic camera operations (\eg, orbit, pan, zoom) as well as dynamic camera trajectories extracted from real videos.

\noindent\textbf{Single-view real-world data for mixed-training.}\quad
The generated paired videos inevitably contain visual artifacts. In our dual-stream mode, the abundant single-view real-world data can be leveraged for mixed-training. First, for the videos with static cameras (object motion only), we duplicate the video and treat it as the target video. In this case, the video model learns to exchange the motion condition from the first stream (with motion condition) into the target stream (without motion condition). Second, for the videos that exhibit both camera and object motion, we feed the condition into the video model as described in~\cref{sec:disentangle} and only supervise the second stream, leaving the loss at the first stream being zero. These two mixed-training strategies expose video models to real-world data with diverse camera and object motion, increasing the robustness and generalizability to various camera and motion configurations, while mitigating artifacts from synthetic data.

\noindent\textbf{Rendered Graphics Data.}\quad
We further incorporate synthetic data from SyncCamMaster~\cite{bai2024syncammaster} to expose our model to more camera diversity.

\subsection{Training and Inference}
\label{sec:training}
We train the DiT using the flow matching loss from~\cref{eq:flow_matching}, and apply two complementary dropout strategies to encourage the model learn the motion causality.

\noindent\textbf{Multi-granularity motion dropout.}\quad
Per~\cref{eq:motion_dropout}, we randomly retain either
$\boldsymbol{\tau}^{\text{act}}$ or $\boldsymbol{\tau}^{\text{pas}}$ to
encourage causal reasoning. We further obtain multi-granularity trajectories by averaging per-pixel trajectories within each patch.
During training, we randomly select the granularity, enabling the model to capture
both fine-grained
pixel control and object-level manipulation.

\noindent\textbf{Occlusion and track dropout.}\quad
For the obtained trajectory, we
randomly mask a subset of it to simulate occlusion and tracking failures that can happen during inference,
improving robustness to missing or unreliable tracks.

\noindent\textbf{Inference.}\quad
At test time, users first specify motion by drawing sparse trajectories (simple curves or strokes on the first image) to indicate the desired
direction and magnitude of movement, along with an optional text prompt and target camera poses $\{C_i\}_{i=1}^{T}$.
We further perform occlusion-aware masking by approximating visibility ordering from the first-frame depth.
The model then jointly denoises both streams, with the second stream being presented to the user.

\vspace{-2mm}
\section{Experiments}
\label{exp}

\subsection{Implementation Details}
We build upon the pretrained Wan2.1-14B~\cite{wan2025wan} and fine-tune only the camera encoder, trajectory encoder, and self-attention layers.
The trajectory embedding dim is 64, and the camera encoder uses 32 channels.
We train the model with 15K iterations on 64 GPUs with a global batch size of 16, using AdamW~\cite{loshchilov2017decoupled} with a learning rate of $3\times10^{-5}$ and weight decay $0.001$.
Trajectory dropout is set to 0.1 and text-conditioning dropout to 0.2.

Following the data curation pipeline in~\cref{sec:data}, we build our training data from large-scale public video datasets, including Panda-70M~\cite{chen2024panda} and Wild-SDG-1M~\cite{huang2025vipe}. From these sources, we collect 76K static-view videos, from which we synthesize 43K paired dynamic-view videos using camera-controlled video-to-video generation, and 3.4K synthetic interaction videos from SyncMaster~\cite{bai2024syncammaster}.
All videos are processed at 480p resolution.
At inference, we sample 35 diffusion steps; generating one video takes approximately 15 minutes on a single A100 GPU. More implementation details are presented in \cref{sec:supp_implement}.

\subsection{Experiment Settings}

\noindent\textbf{Evaluation metrics.}
We evaluate our model across four different aspects:
\emph{Video quality}: PSNR and SSIM against reference videos, and FID~\cite{FID} and FVD~\cite{FVD} for distribution-level similarity.
\emph{Camera accuracy}: rotation and translation errors~\cite{bai2025recammaster, bai2024syncammaster, cameractrl} between reference poses and poses estimated from generated videos using ViPE~\cite{huang2025vipe}; we report \textbf{median} errors across frames to mitigate estimation noise.
\emph{Motion accuracy}: end-point error (EPE)~\cite{chu2025wanmove, geng2024motionprompting}, the $\ell_2$ distance between ground-truth object tracks and predicted tracks extracted with AllTracker; we report the \textbf{median} EPE to reduce the impact of outlier tracks.
\emph{Motion realism}: Physical Commonsense (PC) and Semantic Adherence (SA) from VideoPhy~\cite{bansal2024videophy}, both 5-point scores normalized to $[0,1]$.
All evaluations are conducted at 480p resolution.

\noindent\textbf{Evaluation Datasets.}
We evaluate on three datasets spanning diverse interaction scenarios.
DynPose-100K~\cite{rockwell2025dynamic} is an in-the-wild dataset with highly dynamic camera motion; we manually select 50 videos exhibiting strong viewpoint changes and clear object interactions.
WISA~\cite{wang2025wisa} is a large-scale physical-dynamics dataset; we select 50 videos from categories including collision, deformation, elasticity, liquid, and rigid-body motion.
We further collect 50 real-world cooking videos, featuring complex hand-object interactions.

\begin{figure}[!t]
\centering

\begin{tabular}{c} \includegraphics[width=\linewidth]{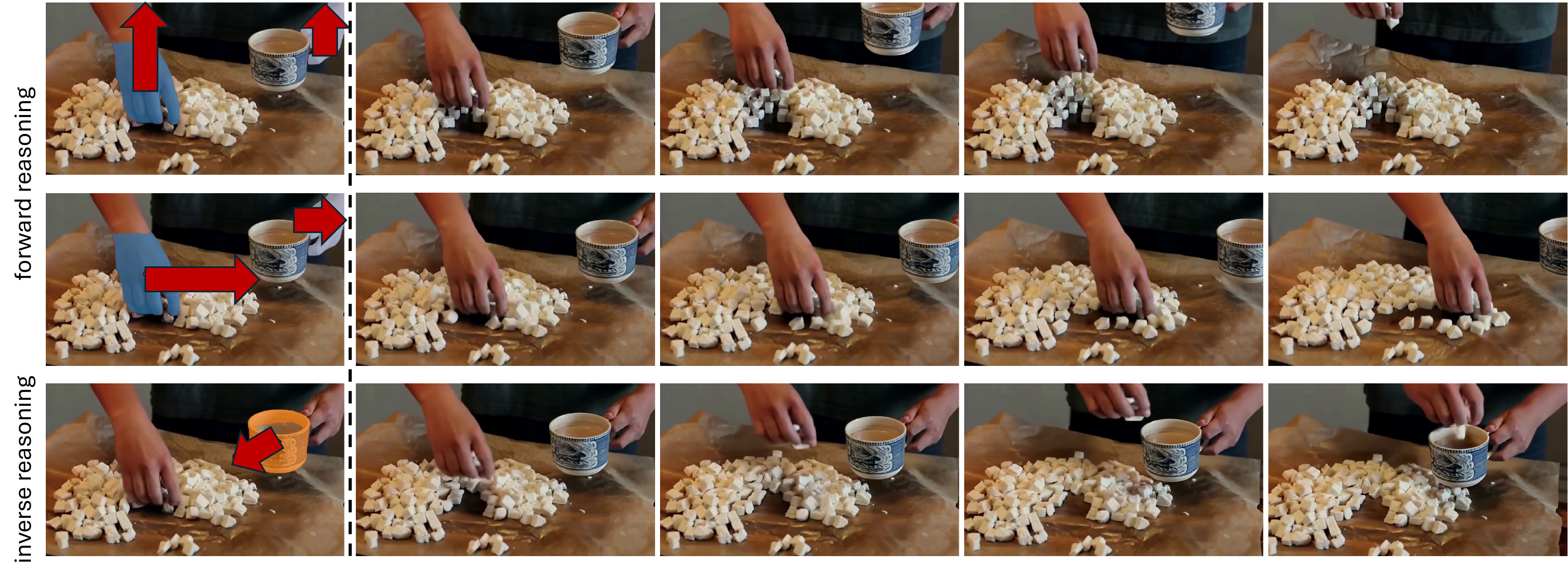} \\
\includegraphics[width=\linewidth]{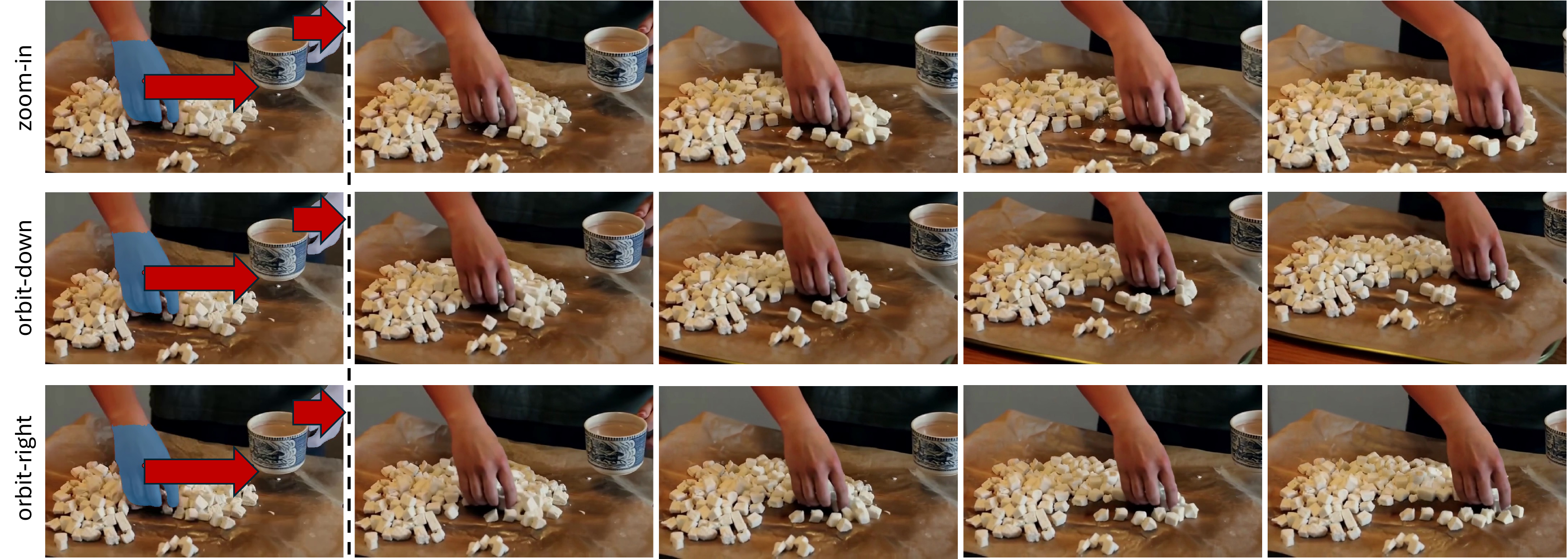} \\
\end{tabular}

\caption{\textbf{Disentangled camera–object control.} \ours enables independent control of object motion and camera viewpoint. Rows 1-3 fix the camera and vary object motion (rows 1-2: forward reasoning; row 3: inverse reasoning), while rows 4-6 fix object motion and vary camera motion.}
\label{fig:control}
\end{figure}

\begin{figure}
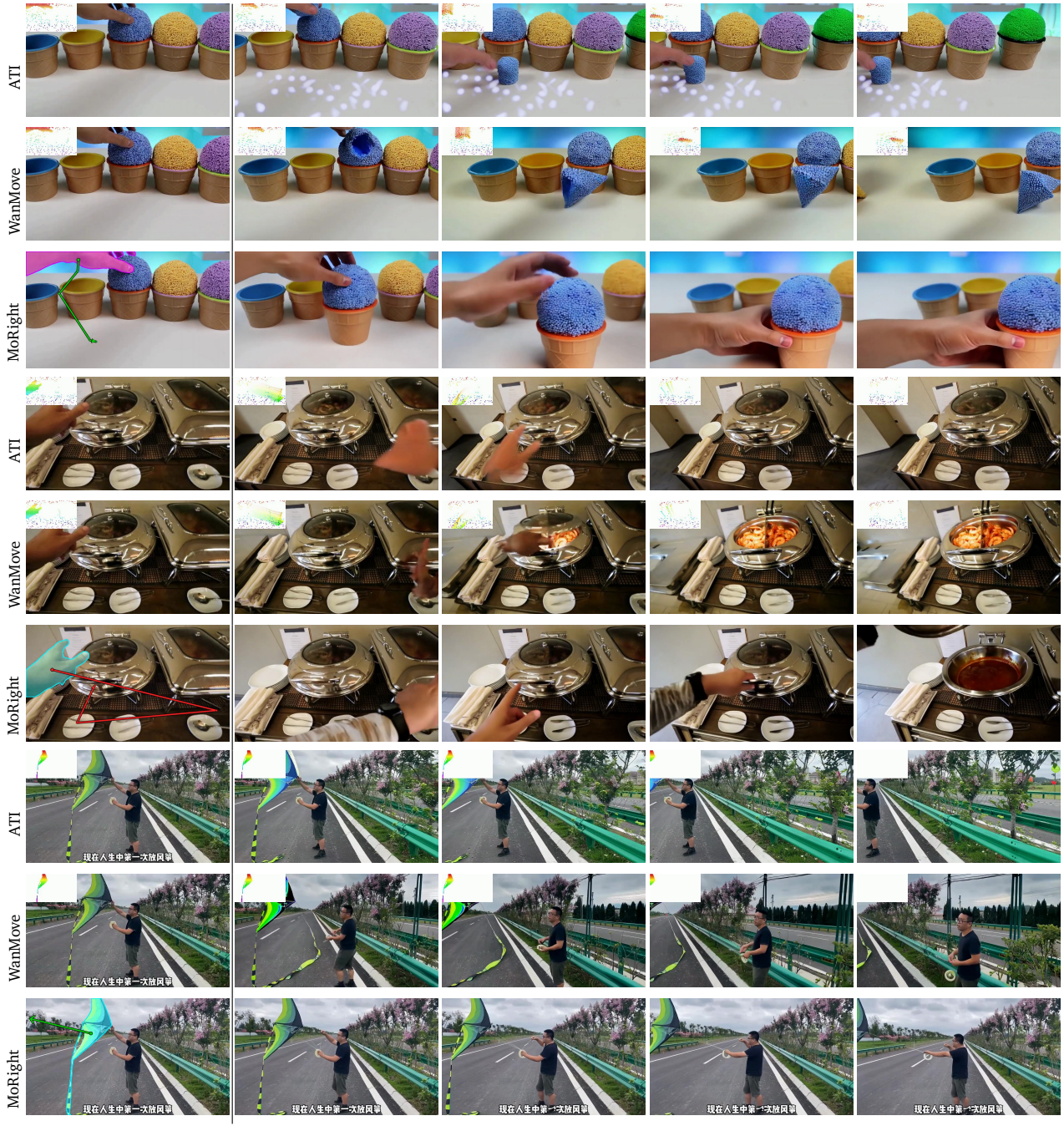

\centering
\setlength{\tabcolsep}{1pt}
\resizebox{\linewidth}{!}{%
\begin{tabular}{@{}c c|cccc@{}}
\raisebox{1.5em}{\rotatebox[origin=c]{90}{\scriptsize ATI}} &
\visrow{comparison}{ati/a6ec1187-ae42-4e9a-8a75-37b6ff9bf4ca} \\[1pt]
\raisebox{1.5em}{\rotatebox[origin=c]{90}{\scriptsize WanMove}} &
\visrow{comparison}{wan_move/a6ec1187-ae42-4e9a-8a75-37b6ff9bf4ca} \\[1pt]
\raisebox{1.5em}{\rotatebox[origin=c]{90}{\scriptsize \ours}} &
\visrow{comparison}{ours/a6ec1187-ae42-4e9a-8a75-37b6ff9bf4ca}{} \\

\raisebox{1.5em}{\rotatebox[origin=c]{90}{\scriptsize ATI}} &
\visrow{comparison}{ati/4e097846-c17e-4cad-bf81-40942bc65883} \\[1pt]
\raisebox{1.5em}{\rotatebox[origin=c]{90}{\scriptsize WanMove}} &
\visrow{comparison}{wan_move/4e097846-c17e-4cad-bf81-40942bc65883} \\[1pt]
\raisebox{1.5em}{\rotatebox[origin=c]{90}{\scriptsize \ours}} &
\visrow{comparison}{ours/4e097846-c17e-4cad-bf81-40942bc65883}{} \\

\raisebox{1.5em}{\rotatebox[origin=c]{90}{\scriptsize ATI}} &
\visrow{comparison}{ati/2ebb7163c3d607663c4a61d49479d4420abbde573877114f7fa0fd15fd4376c4} \\[1pt]
\raisebox{1.5em}{\rotatebox[origin=c]{90}{\scriptsize WanMove}} &
\visrow{comparison}{wan_move/2ebb7163c3d607663c4a61d49479d4420abbde573877114f7fa0fd15fd4376c4} \\[1pt]
\raisebox{1.5em}{\rotatebox[origin=c]{90}{\scriptsize \ours}} &
\visrow{comparison}{ours/2ebb7163c3d607663c4a61d49479d4420abbde573877114f7fa0fd15fd4376c4} \\

\end{tabular}%
}
\caption{
\textbf{Qualitative comparison of ATI~\cite{wang2025ati}, WanMove~\cite{chu2025wanmove}, and \ours} on interactive motion generation with camera control.
All methods use the same input image. ATI and WanMove rely on pixel-aligned per-frame tracks (top-left), which entangle camera and object motion and require privileged future tracks. In contrast, \ours uses only reprojected first-frame tracks. The first two rows show active motion reasoning, and the third row shows passive motion reasoning. Our model disentangles camera and object control and produces more coherent interactions.}
\label{fig:qualitative_compare}
\end{figure}

\begin{table*}[t]
\centering
\caption{\textbf{Controllable video generation on DynPose-100K~\cite{rockwell2025dynamic} and Cooking.}
We compare models with camera and object motion control. Tracking-based methods (MP, ATI, WanMove) require privileged foreground/background tracks, while \ours uses only first-frame reprojected trajectories and camera poses. Despite weaker inputs, \ours achieves comparable visual quality and more accurate motion control. All methods use the Wan2.1-14B backbone; * denotes models reimplemented and trained by us. Best and second-best results are marked in \textbf{bold} and \underline{underline}.}
\label{tab:ctrl_twodsets}

\setlength{\tabcolsep}{3.5pt}
\renewcommand{\arraystretch}{1.12}
\small

\definecolor{ctrlgray}{gray}{0.96}
\adjustbox{width=\textwidth}{
\begin{tabular}{
l c
c c >{\columncolor{ctrlgray}}c >{\columncolor{ctrlgray}}c >{\columncolor{ctrlgray}}c
c c >{\columncolor{ctrlgray}}c >{\columncolor{ctrlgray}}c >{\columncolor{ctrlgray}}c
}
\toprule
& & \multicolumn{5}{c}{\textbf{DynPose-100K}} & \multicolumn{5}{c}{\textbf{Cooking}} \\
\cmidrule(lr){3-7}\cmidrule(lr){8-12}
Method
& Full info
& PSNR $\uparrow$
& SSIM $\uparrow$
& \cellcolor{ctrlgray}Rot $\downarrow$
& \cellcolor{ctrlgray}Trans $\downarrow$
& \cellcolor{ctrlgray}EPE $\downarrow$
& PSNR $\uparrow$
& SSIM $\uparrow$
& \cellcolor{ctrlgray}Rot $\downarrow$
& \cellcolor{ctrlgray}Trans $\downarrow$
& \cellcolor{ctrlgray}EPE $\downarrow$ \\
\midrule
Wan2.1~\cite{wan2025wan} & $\times$
& 11.23 & 0.435 & - & - & --
& 14.23 & 0.527 & - & - & -- \\

Gen3c*~\cite{ren2025gen3c} & $\times$
& \underline{12.45} & \underline{0.507} & 5.46 & \underline{4.09} & -
& 15.37 & \textbf{0.613} & \textbf{1.97} & \textbf{10.03} & - \\

MP*~\cite{geng2024motionprompting} & $\checkmark$
& 11.72 & 0.455 & 6.76 & 6.04 & \underline{7.56}
& 15.68 & 0.564 & 2.50 & 12.24 & \textbf{4.25} \\

ATI~\cite{wang2025ati} & $\checkmark$
& 13.18 & 0.493 & 5.62 & 6.54 & 8.43
& 15.93 & 0.582 & 4.25 & 16.94 & 5.87 \\

WanMove~\cite{chu2025wanmove} & $\checkmark$
& \textbf{13.91} & \textbf{0.521} & \textbf{4.12} & \textbf{3.56} & 8.05
& \underline{16.42} & \underline{0.589} & 2.93 & 13.27 & 5.47 \\

Ours & $\times$
& 12.30 & 0.457 & \underline{4.55} & 4.61 & \textbf{7.64}
& \textbf{16.44} & 0.594 & \underline{2.16} & \underline{10.11} & \underline{4.27} \\
\bottomrule
\end{tabular}
}
\end{table*}
\begin{table*}[t]
\centering
\caption{\textbf{Interactive motion generation on WISA~\cite{wang2025wisa} and Cooking.}
We compare motion-conditioned video generation models on WISA and Cooking, evaluating video quality (FID, FVD) and motion realism (PC, SA). Prior methods require detailed motion captions with full interaction descriptions, while \ours uses only a single active motion description yet achieves comparable quality with stronger physical commonsense reasoning. All methods use the Wan2.1-14B backbone for fair comparison; * denotes models reimplemented and trained by us. Best and second-best results are marked in \textbf{bold} and \underline{underline}.}
\label{tab:interactive_wisa_cooking}

\setlength{\tabcolsep}{4pt}
\renewcommand{\arraystretch}{1.12}
\small

\definecolor{motgray}{gray}{0.96}
\adjustbox{width=0.88\textwidth}{
\begin{tabular}{
l c
c c >{\columncolor{motgray}}c >{\columncolor{motgray}}c
c c >{\columncolor{motgray}}c >{\columncolor{motgray}}c
}
\toprule
& & \multicolumn{4}{c}{\textbf{WISA}} & \multicolumn{4}{c}{\textbf{Cooking}} \\
\cmidrule(lr){3-6}\cmidrule(lr){7-10}
Method
& Full info
& FID $\downarrow$
& FVD $\downarrow$
& \cellcolor{motgray}PC $\uparrow$
& \cellcolor{motgray}SA $\uparrow$
& FID $\downarrow$
& FVD $\downarrow$
& \cellcolor{motgray}PC $\uparrow$
& \cellcolor{motgray}SA $\uparrow$ \\
\midrule
MP*~\cite{geng2024motionprompting} & $\checkmark$
& \underline{57.29} & \underline{975.94} & 0.75 & \underline{0.82}
& \underline{43.49} & \underline{759.53} & 0.87 & \underline{0.89} \\

ATI~\cite{wang2025ati} & $\checkmark$
& 69.80 & 990.82 & 0.75 & \textbf{0.83}
& 55.80 & 881.94 & 0.85 & \textbf{0.90} \\

WanMove~\cite{chu2025wanmove} & $\checkmark$
& 61.34 & 1088.23 & 0.73 & \textbf{0.83}
& 53.51 & 882.90 & 0.84 & 0.87 \\

Ours & $\times$
& \textbf{52.95} & \textbf{876.03} & \textbf{0.76} & \underline{0.82}
& \textbf{39.94} & \textbf{730.46} & \textbf{0.88} & \underline{0.89} \\
\bottomrule
\end{tabular}
}
\end{table*}
\begin{figure}[t]
\centering
\setlength{\tabcolsep}{1.5pt}
\resizebox{\linewidth}{!}{%
\begin{tabular}{@{}c ccccc@{}}
\raisebox{1.5em}{\rotatebox[origin=c]{90}{\scriptsize active}} &
\visrow{active_passive}{double_lift_cloth_1_move_right} \\
\raisebox{1.5em}{\rotatebox[origin=c]{90}{\scriptsize active}} &
\visrow{active_passive}{double_lift_cloth_1} \\
\raisebox{1.5em}{\rotatebox[origin=c]{90}{\scriptsize active}} &
\visrow{active_passive}{double_lift_cloth_1_zoom_in_poke} \\
\midrule
\raisebox{1.5em}{\rotatebox[origin=c]{90}{\scriptsize passive}} &
\visrow{active_passive}{0dad0910edf02720c0ad6ec976c1b21dd6b5fdccfd3786ce5e71b30347de0962_original} \\
\raisebox{1.5em}{\rotatebox[origin=c]{90}{\scriptsize passive}} &
\visrow{active_passive}{0dad0910edf02720c0ad6ec976c1b21dd6b5fdccfd3786ce5e71b30347de0962_straight_right} \\
\raisebox{1.5em}{\rotatebox[origin=c]{90}{\scriptsize passive}} &
\visrow{active_passive}{0dad0910edf02720c0ad6ec976c1b21dd6b5fdccfd3786ce5e71b30347de0962_straight_left} \\
\end{tabular}
}
\caption{\textbf{Causal interaction reasoning.} In the first 3 rows, we provide active motion (\eg, hand movement) as input, and the model infers the resulting passive motion (\eg, cloth movement). In the last 3 rows, we provide passive motion (\eg, ball movement), and the model infers the corresponding active motion (\eg, human movement).}
\label{fig:active_passive}
\end{figure}

\begin{figure}
    \centering
     \includegraphics[width=1.0\linewidth]{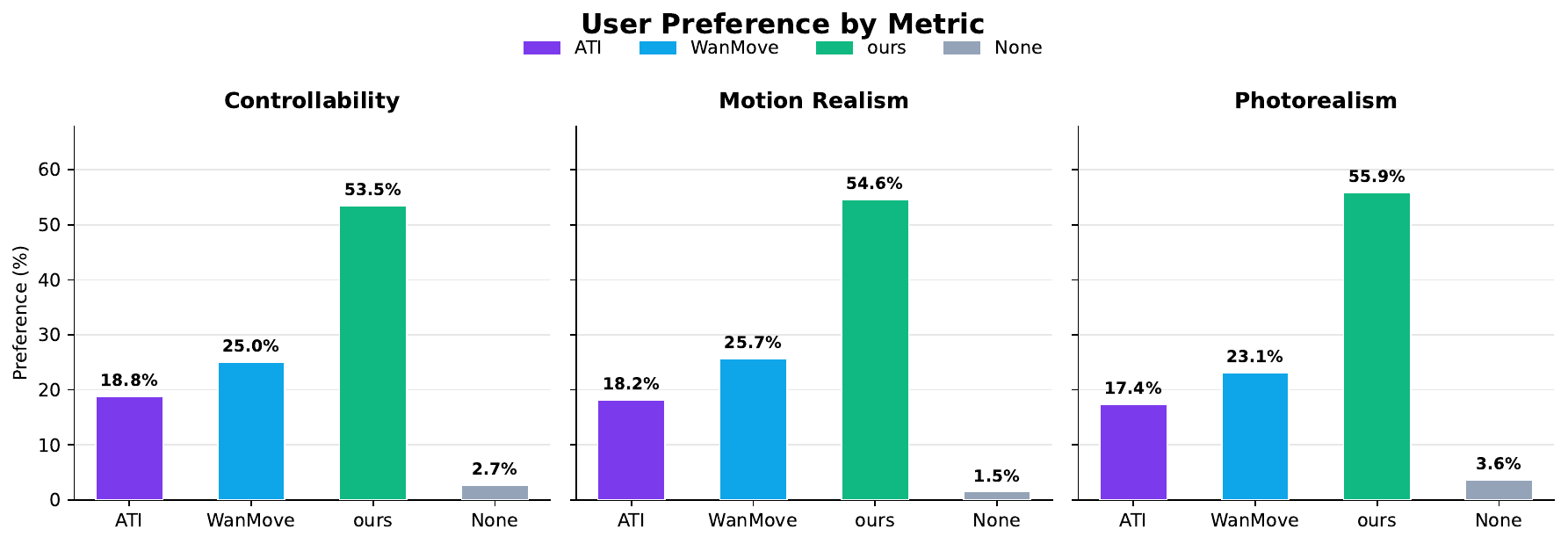}
\caption{\textbf{Human perceptual evaluation.} From 330 responses by 11 participants, our method is preferred across controllability, motion realism, and photorealism, outperforming ATI~\cite{wang2025ati} and WanMove~\cite{chu2025wanmove}, which rely on privileged 3D tracks but lack interaction reasoning.}
\label{fig:user_study_results}
\vspace{-1em}
\end{figure}

\begin{table*}[t]
\centering
\caption{\textbf{Ablation of motion controllability and reasoning on the Cooking benchmark.}
We ablate architectural choices, causal reasoning, hybrid training, and different motion input granularities.
Our full model achieves the best overall performance across photometric quality, controllability, and motion realism, while remaining robust to different motion granularities and input conditions (active and passive).}
\label{tab:ablation}

\setlength{\tabcolsep}{3.5pt}
\renewcommand{\arraystretch}{1.12}
\small

\definecolor{ctrlgray}{gray}{0.96}
\definecolor{motgray}{gray}{0.96}

\adjustbox{width=0.9\textwidth}{
\begin{tabular}{
l
c c c c
>{\columncolor{ctrlgray}}c
>{\columncolor{ctrlgray}}c
>{\columncolor{ctrlgray}}c
>{\columncolor{motgray}}c
>{\columncolor{motgray}}c
}
\toprule
& \multicolumn{4}{c}{\textbf{Photometric}}
& \multicolumn{3}{c}{\cellcolor{ctrlgray}\textbf{Controllability}}
& \multicolumn{2}{c}{\cellcolor{motgray}\textbf{Motion}} \\
\cmidrule(lr){2-5}
\cmidrule(lr){6-8}
\cmidrule(lr){9-10}
\textbf{Setting}
& FID $\downarrow$
& FVD $\downarrow$
& PSNR $\uparrow$
& SSIM $\uparrow$
& \cellcolor{ctrlgray}Rot $\downarrow$
& \cellcolor{ctrlgray}Trans $\downarrow$
& \cellcolor{ctrlgray}EPE $\downarrow$
& \cellcolor{motgray}PC $\uparrow$
& \cellcolor{motgray}SA $\uparrow$ \\
\midrule
cascaded
& 41.74 & \underline{728.80} & 15.98 & 0.569
& 2.69 & 11.50 & 5.05
& 0.87 & 0.89 \\

w/o fixed view
& 51.17 & 997.83 & 14.15 & 0.515
& 3.36 & 14.57 & 14.30
& 0.87 & 0.89 \\

w/o reasoning
& 44.04 & 784.19 & 15.55 & 0.562
& 2.88 & 12.49 & 5.05
& 0.87 & 0.88 \\

w/o mixed training
& 41.94 & 808.96 & 16.29 & 0.583
& 2.22 & 12.80 & \textbf{4.09}
& 0.87 & 0.89 \\

\midrule
ours (coarse)
& \textbf{39.83} & \textbf{725.88} & \textbf{16.45} & \textbf{0.594}
& \underline{2.21} & \underline{10.98} & 4.37
& \textbf{0.88} & 0.88 \\

ours (passive)
& 44.20 & 838.67 & 15.99 & 0.588
& \underline{2.21} & 11.04 & 7.27
& 0.87 & 0.88 \\

ours (active)
& \underline{39.94} & 730.46 & \underline{16.44} & \underline{0.594}
& \textbf{2.16} & \textbf{10.11} & \underline{4.27}
& \textbf{0.88} & 0.89 \\

\bottomrule
\end{tabular}
}
\end{table*}
\begin{figure}
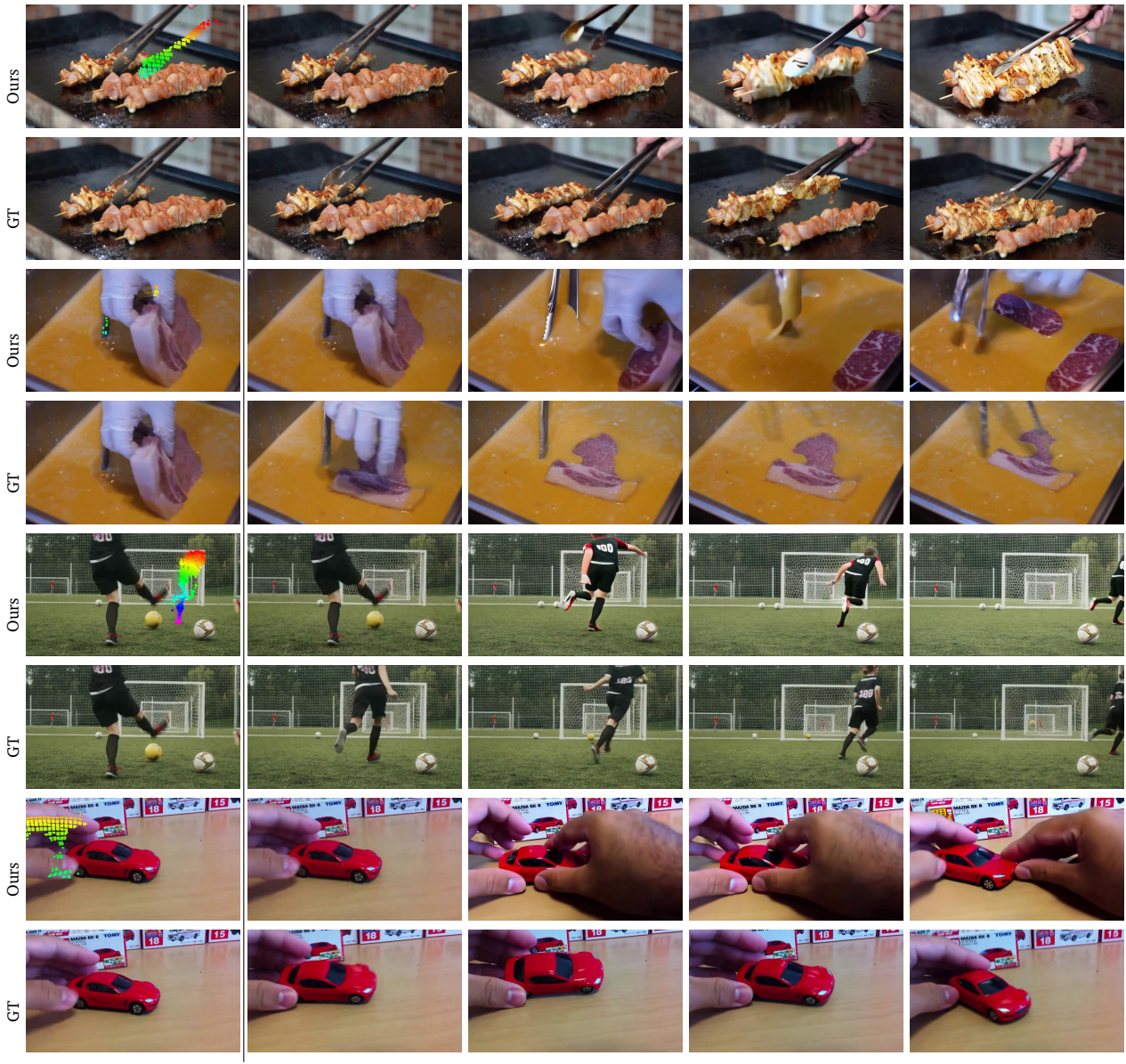

\centering
\setlength{\tabcolsep}{1.5pt}
\resizebox{\linewidth}{!}{%
\begin{tabular}{@{}c c|cccc@{}}
\raisebox{1.5em}{\rotatebox[origin=c]{90}{\scriptsize Ours}} &
\visrow{failures}{0f3eec08-4727-4e16-9d72-660bdf825803} \\
\raisebox{1.5em}{\rotatebox[origin=c]{90}{\scriptsize GT}} &
\visrow{failures}{0f3eec08-4727-4e16-9d72-660bdf825803_gt_video} \\

\raisebox{1.5em}{\rotatebox[origin=c]{90}{\scriptsize Ours}} &
\visrow{failures}{01e63d8c-f5d8-421b-8d8d-e3baa02472d1} \\
\raisebox{1.5em}{\rotatebox[origin=c]{90}{\scriptsize GT}} &
\visrow{failures}{01e63d8c-f5d8-421b-8d8d-e3baa02472d1_gt_video} \\

\raisebox{1.5em}{\rotatebox[origin=c]{90}{\scriptsize Ours}} &
\visrow{failures}{a00fa6501338b10a0a72c94bf4d971571f5f44960245b0863fe319dfe8ce9ba1} \\
\raisebox{1.5em}{\rotatebox[origin=c]{90}{\scriptsize GT}} &
\visrow{failures}{a00fa6501338b10a0a72c94bf4d971571f5f44960245b0863fe319dfe8ce9ba1_gt_video} \\

\raisebox{1.5em}{\rotatebox[origin=c]{90}{\scriptsize Ours}} &
\visrow{failures}{d79eb4e5-bdc3-4cf3-bc7e-f1ea4eda320d} \\
\raisebox{1.5em}{\rotatebox[origin=c]{90}{\scriptsize GT}} &
\visrow{failures}{d79eb4e5-bdc3-4cf3-bc7e-f1ea4eda320d_gt_video} \\

\end{tabular}
}
\caption{\textbf{Limitation analysis.} Input tracks are overlaid on the first frame as in previous figures. (1) Incorrect interaction reasoning may lead to implausible outcomes (two kabobs merging). (2) Unnatural motion can occur when input tracks become temporally sparse due to occlusion (hand example). (3) Physically unrealistic dynamics may appear, such as objects disappearing during motion (soccer ball). (4) Hallucinated content may emerge in later frames (extra hand).}
\label{fig:supp_limitation}
\end{figure}

\subsection{Disentangled Camera-Object Motion Control}
\label{sec:ctrl_generation}

Existing works on controllable video generation typically focus on either camera or object motion in isolation. Motion-conditioned baselines~\cite{geng2024motionprompting,wang2025ati,chu2025wanmove} typically receive privileged signals of both foreground and background tracks of all the pixel trajectories for control, while our method only uses reprojected trajectories defined on the canonical frame, without access to future-frame pixel trajectories. We compare our method with state-of-the-art baselines to evaluate the quality in motion control and camera control. While challenging, this evaluation allows us to demonstrate our model's unique ability to generate faithful motion from disentangled controls—a task that is fundamentally more difficult than baselines.

\noindent\textbf{Baselines and setup.}
We compare with several controllable video generation methods.
Wan2.1~\cite{wan2025wan} is our base model without motion control.
Gen3C~\cite{ren2025gen3c} only supports camera control. Recent state-of-the-art motion-conditioned models (Motion Prompting (MP)~\cite{geng2024motionprompting}, ATI~\cite{wang2025ati} and
WanMove~\cite{chu2025wanmove} all take dense pixel tracks as input.
For a fair comparison, we retrain Gen3C and MP in our setup, and all methods share the same Wan2.1-14B backbone. Evaluation is conducted on DynPose-100K~\cite{rockwell2025dynamic} and Cooking dataset. Motion accuracy is evaluated in future-frame pixel space via EPE for all methods.

\noindent\textbf{Results.}
Quantitative results are provided in \cref{tab:ctrl_twodsets}, with  controllable results in \cref{fig:control}, qualitative comparisons in \cref{fig:qualitative_compare}.
On DynPose-100K~\cite{rockwell2025dynamic}, WanMove~\cite{chu2025wanmove} achieves the best overall numbers.
Our method is slightly behind under highly dynamic camera motion, where errors in camera pose estimation and trajectory reprojection can degrade the input control signals.
Nevertheless, \ours attains comparable controllability to methods that rely on privileged future-frame tracking information and achieves the best EPE for object motion accuracy.
We observe that ATI~\cite{wang2025ati} and WanMove~\cite{chu2025wanmove}, which couple camera and object motion in a single tracking signal, tend to favor the dominant motion mode in highly dynamic settings---sometimes sacrificing camera accuracy or object tracking fidelity.
On the Cooking benchmark, our method achieves the best overall performance in both visual quality and motion control accuracy. More controllable generation results are shown in \cref{fig:supp_control}, and qualitative comparisons are provided in \cref{fig:supp_qualitative_compare}.

\subsection{Motion Causality Modeling}
\label{sec:exp_casual}

\noindent\textbf{Setup.}
We evaluate the causality of modeling motion on WISA~\cite{wang2025wisa} and Cooking, measuring both generation quality (FID, FVD) and motion realism (PC, SA).
We compare with MP*~\cite{geng2024motionprompting}, ATI~\cite{wang2025ati}, and WanMove~\cite{chu2025wanmove}, all following the same input protocol as \cref{sec:ctrl_generation}.
Every model receives \emph{active motion} representing the user-specified action (e.g., a hand pushing an object); the goal is to generate plausible interaction outcomes.
Baseline methods use their original prompts containing both motion descriptions and expected consequences.
Our model receives only the active motion description, without specifying passive outcomes, and must infer the resulting interactions.

\noindent\textbf{Results.}
Quantitative results are reported in \cref{tab:interactive_wisa_cooking}.
\ours achieves the highest PC score on WISA, indicating strong physical commonsense, and the best video quality (FID, FVD) on both datasets.
For SA, we rewrite the input prompt to remove passive motion descriptions and avoid information leakage; consequently, our score is slightly lower than methods that use full prompts containing both actions and outcomes, yet remains comparable.
This confirms that \ours's generations stay semantically aligned with intended outcomes while demonstrating genuine causal motion reasoning rather than relying on prompt-supplied answers.
Qualitative examples of both reasoning modes---\emph{forward} (action $\rightarrow$ reaction) and \emph{inverse} (reaction $\rightarrow$ action)---are shown in \cref{fig:active_passive}. Our model generates plausible reactions when providing the active motion, and can reason the meaningful active motion when providing passive motion. More qualitative visualizations are shown in \cref{fig:supp_reasoning}.

\subsection{Human Perceptual Evaluation}
\label{sec:human_eval}

In addition to objective metrics, we conduct a human perceptual study to evaluate generation quality. ATI~\cite{wang2025ati} and WanMove~\cite{chu2025wanmove} rely on pixel-aligned per-frame tracks projected from privileged 3D trajectories, including both foreground/background and full interaction (active and passive) motion. In contrast, our method uses only first-frame active trajectories, requiring the model to infer interactions without privileged information.

We randomly sample 30 examples from the combined test datasets. For each example, videos from different methods are presented in randomized order to avoid positional bias. Participants evaluate results based on three criteria: \emph{Controllability} (alignment with input object and camera motion), \emph{Motion Realism} (physical plausibility of interactions), and \emph{Photorealism} (visual quality). For each criterion, participants select the best result, with ties and a \textit{None} option allowed. After filtering unreliable submissions, we collect responses from 11 participants, yielding 330 evaluations per criterion (the evaluation interface is shown in \cref{fig:user_study_interface}).

As shown in \cref{fig:user_study_results}, our method is preferred in the majority of cases, achieving 53.5\%, 54.6\%, and 55.9\% for controllability, motion realism, and photorealism, respectively. This outperforms ATI~\cite{wang2025ati} (18.8\%, 18.2\%, 17.4\%) and WanMove~\cite{chu2025wanmove} (25.0\%, 25.7\%, 23.1\%). Despite access to privileged 3D trajectories, baseline methods lack explicit interaction reasoning and entangle camera and object motion, leading to inferior performance. In contrast, our disentangled formulation enables more controllable and realistic video generation.

\subsection{Ablation Studies}

We ablate model design, training strategies, and input conditions on the Cooking dataset in~\cref{tab:ablation}.

\noindent\textbf{Model design and training.}
\emph{Cascaded pipeline} (row~1): a naive solution for disentangling camera-object motion is to first generate motion-controlled video under a static camera, followed by a Gen3C-style camera controller to move the camera. However, this approach introduces error accumulation between two stages, yielding larger control errors.
\emph{W/o fixed-view branch} (row~2): we only train with dynamic camera views and jointly encode the reprojected tracks and camera embeddings, removing the canonical-view anchor.  The model struggles to disentangle camera and object motion, resulting in significantly worse camera and tracking accuracy.
\emph{W/o motion reasoning} (row~3): we disable active/passive decomposition during training. This approach increases FID/FVD and reduces PC, indicating degraded interaction quality.
\emph{W/o mixed supervision} (row~4): We only train the model on paired data. It slightly degrades camera accuracy, as the paired subset contains limited camera motion diversity.

\noindent\textbf{Input conditions.}
We vary the motion input configuration to evaluate the robustness of our models. Specifically, we evaluate with coarse segment-level trajectories vs.\ fine-grained pixel tracks, as well as active vs.\ passive motion inputs.
Performance remains stable across all settings, confirming that \ours flexibly handles different motion granularities and types while maintaining strong controllability and causal reasoning capability.

\subsection{Limitation Analysis}
\label{sec:supp_limitation}

Despite promising results, our method still exhibits several limitations, as illustrated in \cref{fig:supp_limitation}. First, the model may produce incorrect interaction reasoning, leading to implausible outcomes such as two kabobs merging into a single object. Second, unnatural motion can occur when the input trajectories become temporally sparse due to occlusion, making it difficult for the model to reliably infer the intended motion (e.g., the hand example). Third, the generated motion may violate physical consistency, such as objects disappearing during interaction (soccer example). Fourth, the model may occasionally hallucinate new content in later frames, such as an extra hand appearing during generation.

In addition, our method has difficulty modeling very complex or fast camera motion (\eg, drastic egomotion). Our camera control is designed for common smooth camera trajectories, and when the input camera motion changes drastically, the predicted interaction dynamics may degrade.

\section{Conclusion}

We present \ours, a unified framework for controllable and interaction-aware video generation. \ours addresses two key limitations of prior motion-controlled methods: (1) entangled camera and object motion, resolved through a dual-stream design that enables independent control of object trajectories and camera viewpoints; and (2) limited causal reasoning, addressed by decomposing motion into active (user-driven) and passive (consequence) components to learn action--response dynamics. At inference, \ours supports both forward prediction—generating scene outcomes from active motion—and inverse reasoning—inferring actions from desired passive results. Experiments on DynPose-100K, WISA, and Cooking show strong performance in generation quality, motion control, and interaction awareness, establishing \ours as a step toward more interactive and physically grounded video generation.

\section*{Acknowledgement}
We would like to thank Jiahui Huang, Zian Wang, Xiao Fu, and Chen-Hsuan Lin for their help and support with video data processing, data generation, and infrastructure.

\renewcommand{\figdir}{supp_src_figs}
\appendix
\clearpage

\title{Appendix}
\author{}
\maketitle
\vspace{-4em}

\section{Implementation Details}
\label{sec:supp_implement}

\subsection{Network Architecture}
Our model builds on the Wan2.1 I2V-14B~\cite{wan2025wan}. We first encode the two-view videos and then concatenate the tokens along the temporal dimension before feeding them into the model. The two streams share the same spatial RoPE~\cite{su2024roformer} embeddings but use different temporal indices. The object tracking condition, represented as a trajectory map, is encoded by a lightweight temporal encoder with RMSNorm, SiLU~\cite{elfwing2018sigmoid}, and two \(3\times1\times1\) Conv3D layers that downsample the temporal dimension by \(4\times\) to match the Wan latent resolution. For camera motion control, we follow Gen3C~\cite{ren2025gen3c} by warping the first frame with the camera trajectory and encoding it with the VAE, producing features in the same latent space and resolution.

Both camera and tracking features are linearly projected to the Wan hidden dimension (5120) and added to the video tokens before the self-attention layer of each Wan2.1 transformer block. During training, we only train the lightweight temporal encoder and the self-attention layers together with the camera and tracking encoders in each block, and freeze other parts of the network.

\subsection{Training Data Curation}
In training data curation, we need to identify active and passive object and its motion in a given video. We first identify those objects by querying Qwen3-VL~\cite{Qwen3-VL} and use SAM2~\cite{ravi2024sam2} for video object segmentation. The system prompt to Qwen3-VL~\cite{Qwen3-VL} is shown in \cref{fig:supp_dynamic_prompt}. We further use Qwen3 to rewrite video captions by decomposing object motion into active or passive descriptions, ensuring that each rewritten caption contains only one type of motion. During training, the original caption and the rewritten caption are randomly sampled with equal probability, encouraging the model to infer plausible interaction consequences. To generate paired multi-view data, we select videos from our collected Internet videos with nearly static cameras using the camera poses provided by ViPE~\cite{huang2025vipe}, requiring a maximum rotation of $0.5^\circ$ and translation of $5\,\text{mm}$.

\begin{figure}[h]
\begin{tcolorbox}[colback=white]
\footnotesize
\textbf{System:} You are a video understanding specialist.  
Follow all rules exactly and output only valid JSON.
\begin{itemize}
    \item \texttt{active\_dynamic}: objects that move by their own power or actuation (e.g., person, hand, animal, robot, drone, car, bus, train, boat, ship, airplane, motorcycle).
    \item \texttt{passive\_dynamic}: objects that move only because of other objects or forces. If a person/hand is visible manipulating an item, that item is \texttt{passive\_dynamic}.
\end{itemize}

\textbf{User:} You are given a video. Identify all objects that actually move:

\begin{itemize}
    \item \texttt{active\_dynamic}: objects moving by their own power (or actuation).
    \item \texttt{passive\_dynamic}: objects that move only because other objects or natural forces cause them to move.
\end{itemize}
\end{tcolorbox}
\vspace{-1em}
\caption{Prompt used for active and passive object identification for Qwen3~\cite{Qwen3-VL} in data curation pipeline.}
\label{fig:supp_dynamic_prompt}
\end{figure}

\subsection{Training}

During training, we apply several data augmentations to improve robustness.
For each sample, we randomly simplify the input trajectories with probability 0.5, where tracks are averaged per object such that all pixels of the same object share a single trajectory.
To encourage the model to reason about motion causality, we randomly provide \emph{active} or \emph{passive} motion tracks with probabilities 0.8 and 0.2, respectively.
We further apply motion dropout by randomly dropping visible tracks with probability 0.2 to simulate occlusion and tracking errors commonly observed in off-the-shelf trackers at inference time.
In addition, we randomly truncate tracks after a sampled middle frame to simulate partial observations of motion.
During training, we randomly sample between 500 and 2000 tracks per iteration.
At inference time, we fix the number of input tracks to 1500 for all experiments to ensure consistent evaluation.
Finally, since multi-view supervision is critical for learning camera--object disentanglement, we control the sampling ratio of multi-view and single-view training samples, ensuring that single-view data is sampled at a lower rate to prevent the model from overfitting to single-view motion patterns.

\subsection{Inference}

\begin{figure}
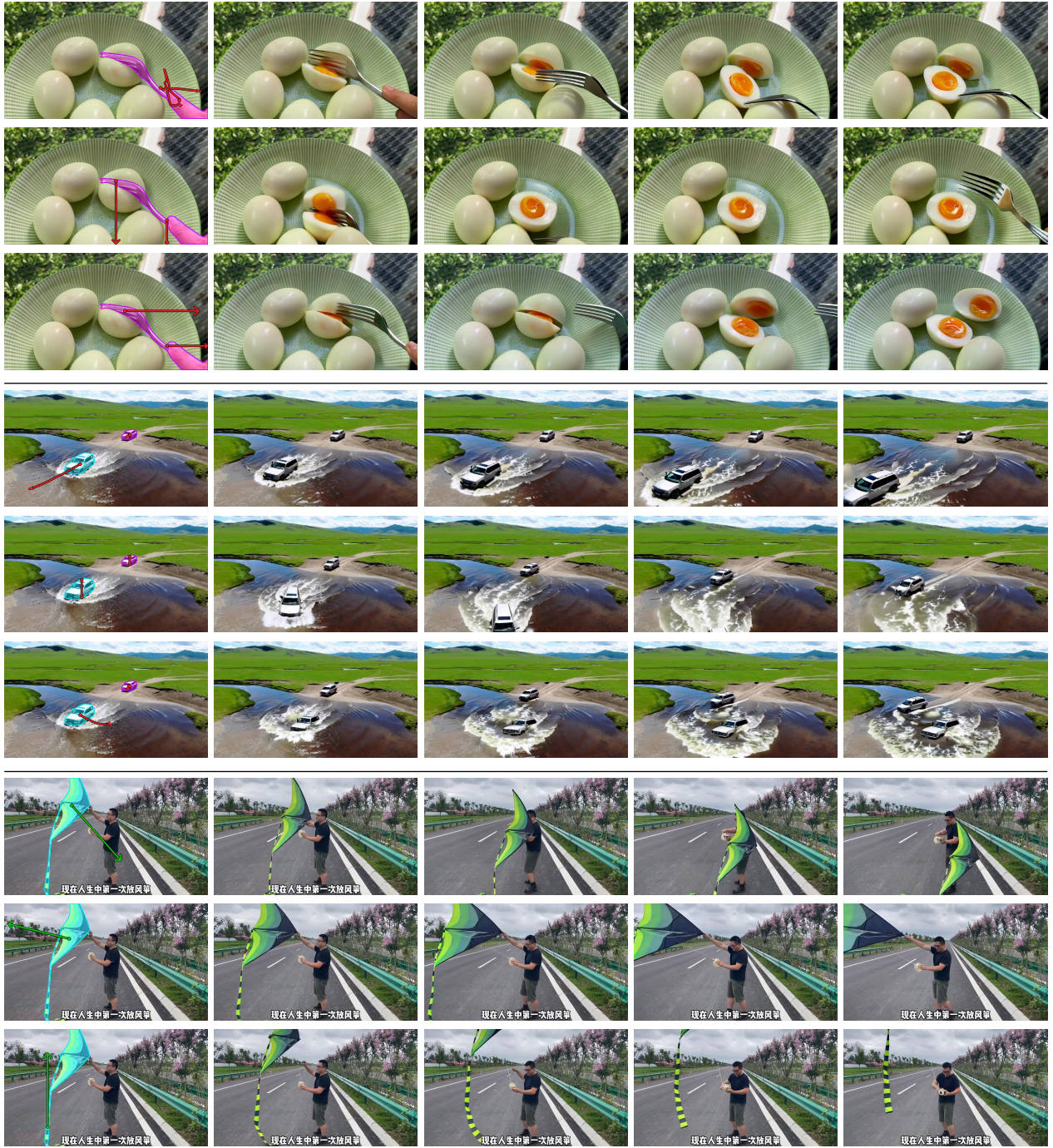

\centering
\setlength{\tabcolsep}{1.5pt}
\resizebox{\linewidth}{!}{%
\begin{tabular}{@{}ccccc@{}}

\visrow{interactive_motion}{fa6b5178-71e9-453b-b6b8-153b6216e903} \\
\visrow{interactive_motion}{fa6b5178-71e9-453b-b6b8-153b6216e903_move_down} \\
\visrow{interactive_motion}{fa6b5178-71e9-453b-b6b8-153b6216e903_move_right} \\

\midrule

\visrow{interactive_motion}{154ccbd9-507f-4596-8481-57c92c223b9e_original} \\
\visrow{interactive_motion}{154ccbd9-507f-4596-8481-57c92c223b9e_move_down} \\
\visrow{interactive_motion}{154ccbd9-507f-4596-8481-57c92c223b9e_push} \\

\midrule

\visrow{interactive_motion}{2ebb7163c3d607663c4a61d49479d4420abbde573877114f7fa0fd15fd4376c4_straight} \\
\visrow{interactive_motion}{2ebb7163c3d607663c4a61d49479d4420abbde573877114f7fa0fd15fd4376c4_straight_left} \\
\visrow{interactive_motion}{2ebb7163c3d607663c4a61d49479d4420abbde573877114f7fa0fd15fd4376c4_straight_up} \\

\end{tabular}
}
\caption{\textbf{Causal interaction reasoning.} Input tracks are shown in color and overlaid on the generated static reference-view video. The tracks represent user actions (active) or passive trajectories. Given these inputs, our model either predicts plausible consequences (forward reasoning) or recovers feasible driving actions that produce the desired outcomes (inverse reasoning, last row).}
\label{fig:supp_reasoning}
\end{figure}

At inference time, users can freely select objects in the first frame and specify their motion trajectories.
Motion control can be provided either in a coarse manner, where the entire object moves with a shared trajectory, or in a fine-grained manner using sparse point tracks.
To facilitate interaction-driven editing, we also provide several simple motion primitives for hand interactions, such as push, pull (along a specified direction), and reach (toward a target location).
For passive motion control, users can define arbitrary 2D trajectories; \cref{fig:supp_reasoning} illustrates an example using straight-line trajectories with different directions for inverse reasoning.

\begin{figure}
    \centering
     \includegraphics[width=0.9\linewidth]{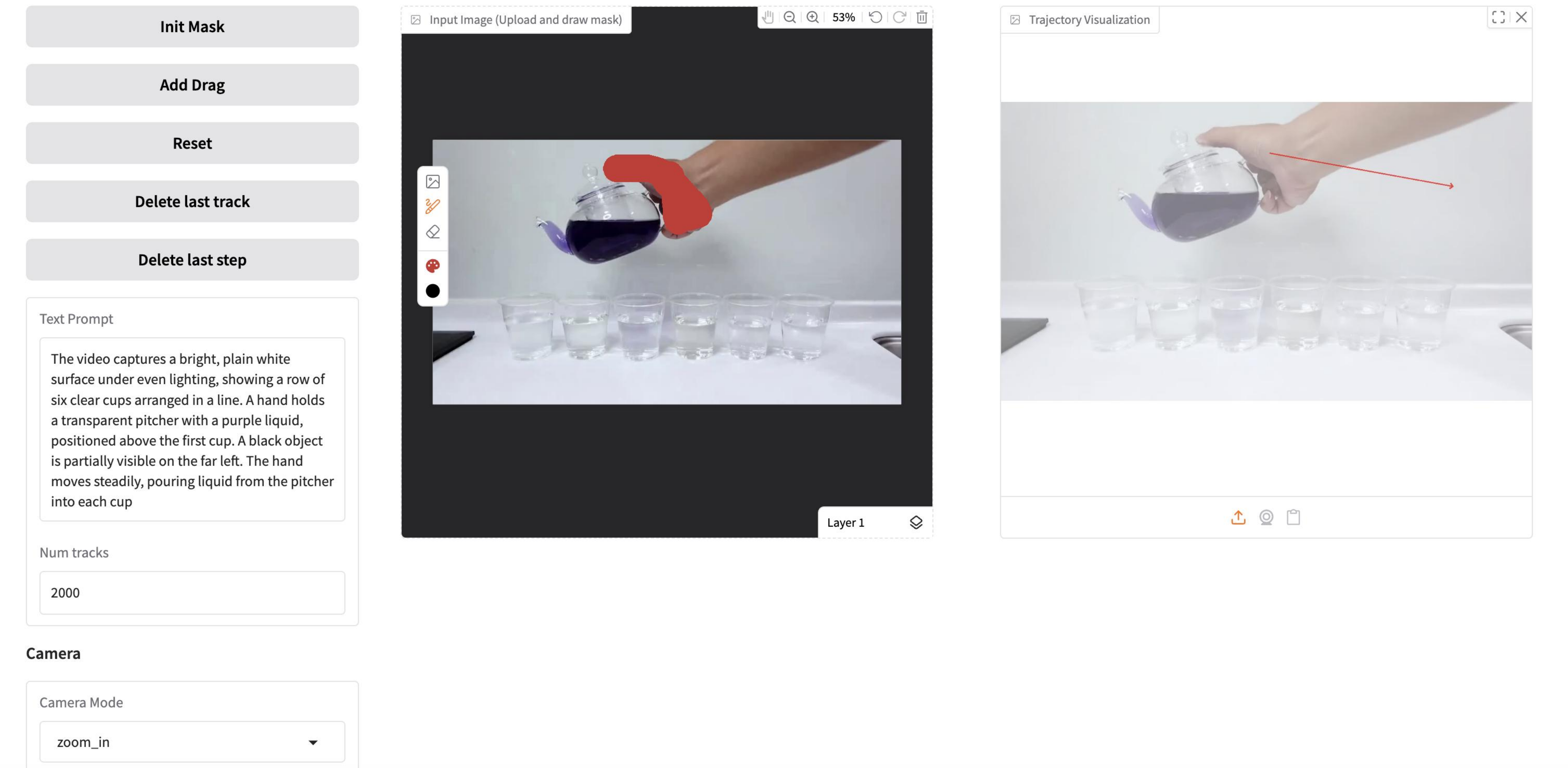} 
\caption{\textbf{Interactive demo interface.} Our system enables users to control both object and camera motion from a single image. Users draw trajectories on the first frame to specify object motion (active or passive), either by moving a selected region using keypoint trajectories or by defining fine-grained motion paths for detailed control.}
\label{fig:gradio_interface}
\end{figure}

To enable flexible control, we implement an interactive GUI as shown in \cref{fig:gradio_interface}.
Starting from a single input image, users draw motion trajectories directly on the first frame while specifying camera motion independently through a sequence of camera poses (the first frame is treated as the identity pose).
The interface supports trajectory visualization across time steps and occlusion checking using the first-frame depth estimated by MoGe~\cite{wang2024moge}, enabling intuitive editing of object dynamics and camera viewpoints during generation.

\subsection{Evaluation}

The \textbf{Cooking Benchmark} is constructed from real-world cooking videos collected from YouTube that contain rich hand–object interactions. These scenes involve diverse manipulation behaviors such as pushing, cutting, and picking, making them a suitable testbed for evaluating interactive motion reasoning. The benchmark contains 50 video clips covering a variety of kitchen environments and object interactions.

For motion quality evaluation, we adopt \textbf{Physical Commonsense (PC)} and \textbf{Semantic Adherence (SA)}, from~\cite{bansal2024videophy, bansal2025videophy}. PC measures whether the generated video follows real-world physical behaviors. SA evaluates whether the generated video remains semantically consistent with the input text prompt. For SA, we use the original caption of each video as the evaluation prompt. We use the automatic evaluation rater provided to score the generated videos and report the resulting normalized scores across different methods.

\begin{figure}
    \centering
     \includegraphics[width=0.9\linewidth]{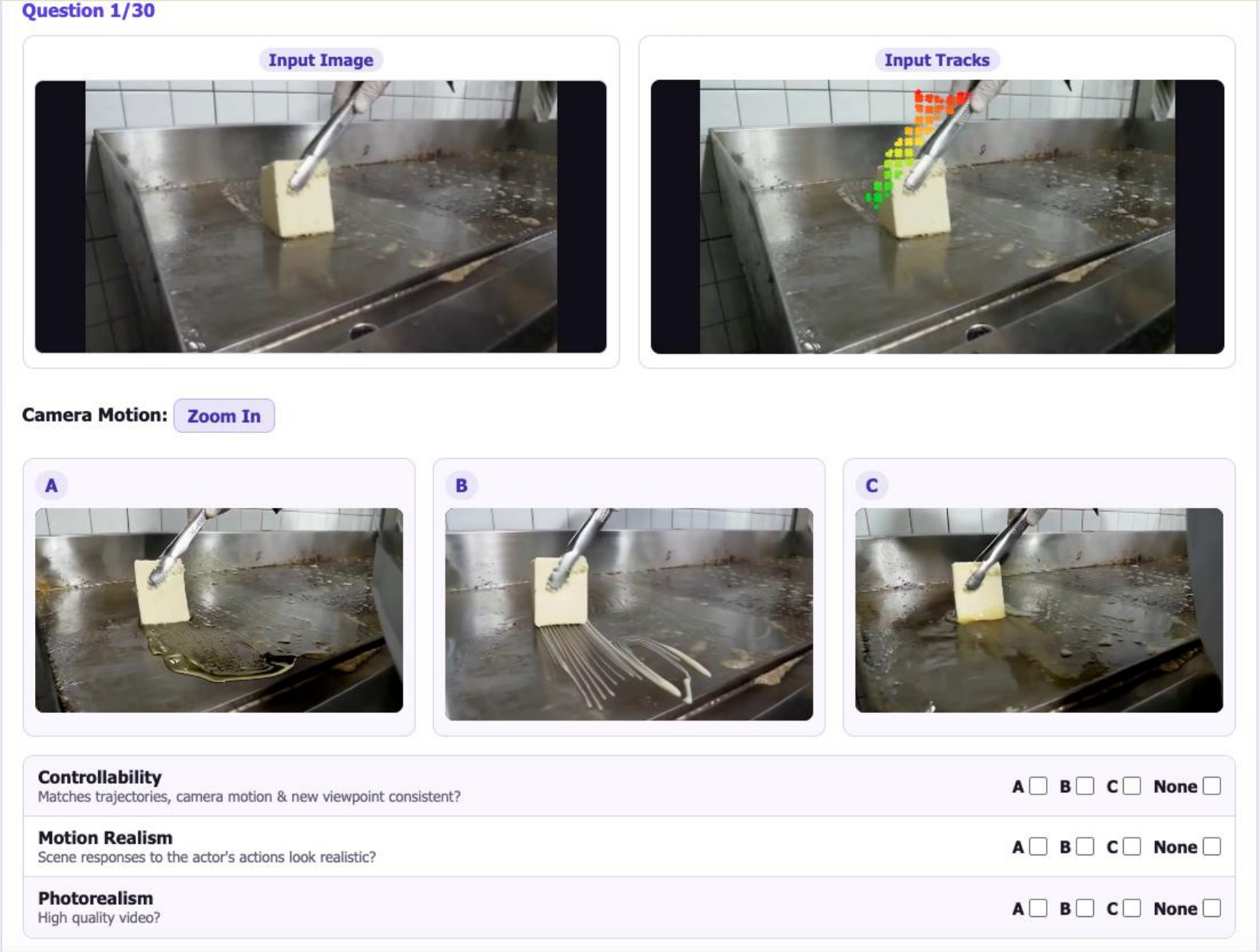} 
\caption{\textbf{Human perceptual evaluation interface.} Given an input image, object trajectories, and a target camera motion, participants evaluate generated videos under three criteria: \emph{Controllability} (matching object tracks and camera motion), \emph{Motion Realism} (physically plausible interactions and scene responses), and \emph{Photorealism} (overall visual quality). For each criterion, participants select the best video (multiple selections allowed for ties, or \textit{None} if none satisfy it). The study contains 30 randomly selected video sets with shuffled candidate order.}
\label{fig:user_study_interface}
\end{figure}

For human perceptual evaluation, the interface layout is shown in \cref{fig:user_study_interface}.

\begin{figure}
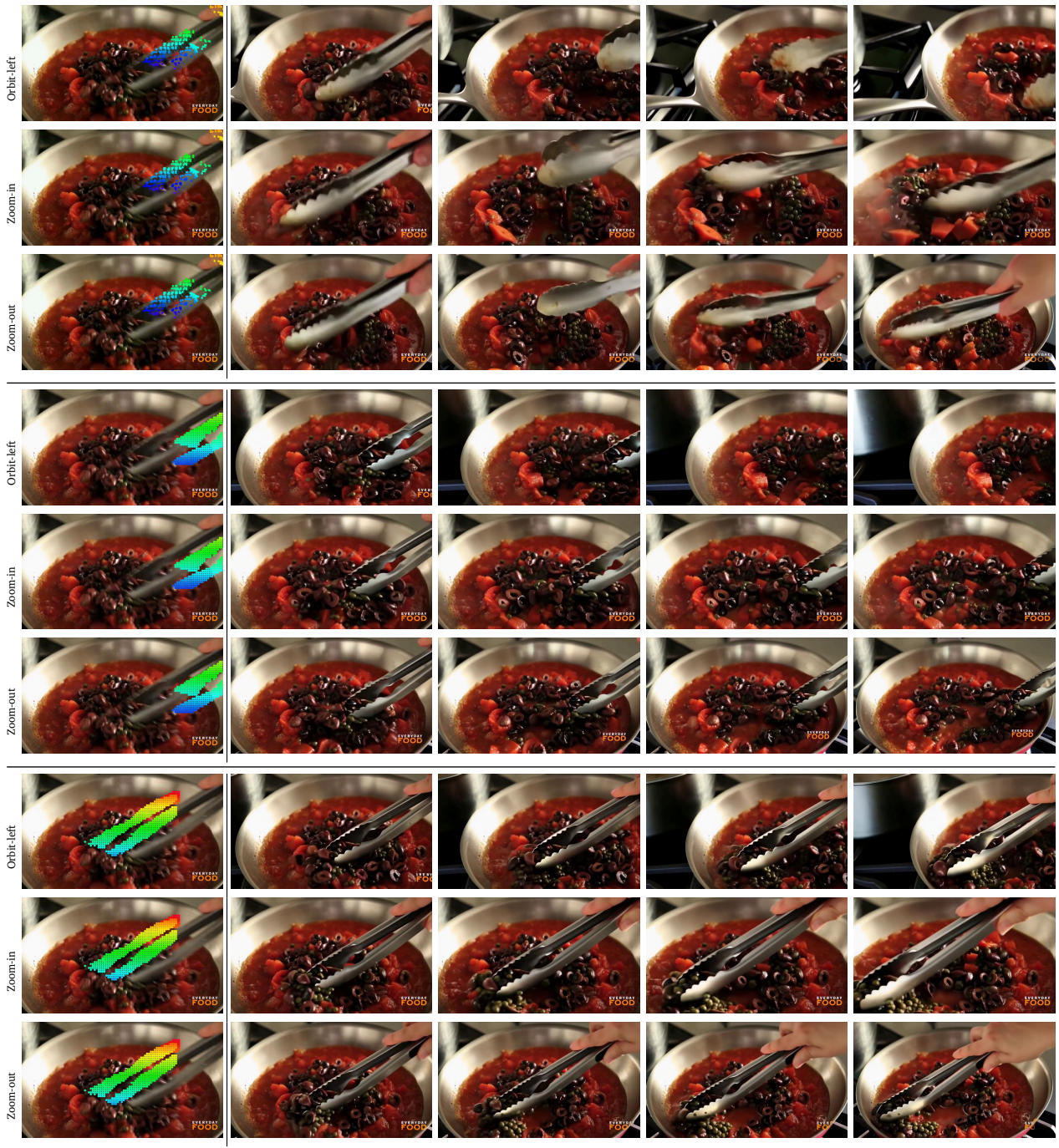

\centering
\setlength{\tabcolsep}{1.5pt}
\resizebox{\linewidth}{!}{%
\begin{tabular}{@{}c c|cccc@{}}
\raisebox{1.7em}{\rotatebox[origin=c]{90}{\tiny Orbit-left}} &
\visrow{control_gen}{6d82d07b-755f-469c-a9fb-ea7a0d60116d_orbit_left_original} \\
\raisebox{1.7em}{\rotatebox[origin=c]{90}{\tiny Zoom-in}} &
\visrow{control_gen}{6d82d07b-755f-469c-a9fb-ea7a0d60116d_zoom_in_original} \\
\raisebox{1.7em}{\rotatebox[origin=c]{90}{\tiny Zoom-out}} &
\visrow{control_gen}{6d82d07b-755f-469c-a9fb-ea7a0d60116d_zoom_out_original} \\

\midrule

\raisebox{1.7em}{\rotatebox[origin=c]{90}{\tiny Orbit-left}} &
\visrow{control_gen}{6d82d07b-755f-469c-a9fb-ea7a0d60116d_orbit_left_pull} \\
\raisebox{1.7em}{\rotatebox[origin=c]{90}{\tiny Zoom-in}} &
\visrow{control_gen}{6d82d07b-755f-469c-a9fb-ea7a0d60116d_zoom_in_pull} \\
\raisebox{1.7em}{\rotatebox[origin=c]{90}{\tiny Zoom-out}} &
\visrow{control_gen}{6d82d07b-755f-469c-a9fb-ea7a0d60116d_zoom_out_pull} \\

\midrule

\raisebox{1.7em}{\rotatebox[origin=c]{90}{\tiny Orbit-left}} &
\visrow{control_gen}{6d82d07b-755f-469c-a9fb-ea7a0d60116d_orbit_left_push} \\
\raisebox{1.7em}{\rotatebox[origin=c]{90}{\tiny Zoom-in}} &
\visrow{control_gen}{6d82d07b-755f-469c-a9fb-ea7a0d60116d_zoom_in_push} \\
\raisebox{1.7em}{\rotatebox[origin=c]{90}{\tiny Zoom-out}} &
\visrow{control_gen}{6d82d07b-755f-469c-a9fb-ea7a0d60116d_zoom_out_push} \\

\end{tabular}
}
\caption{\textbf{Additional controllable generation-1.} Object motion trajectories are overlaid on the input image. For each video, we show different camera and object motion control. Each group shares the same object motion but uses different camera motions. Minor variations under the same object motion but different camera motions arise from the stochastic nature of interaction generation.}
\label{fig:supp_control}
\end{figure}

\begin{figure}
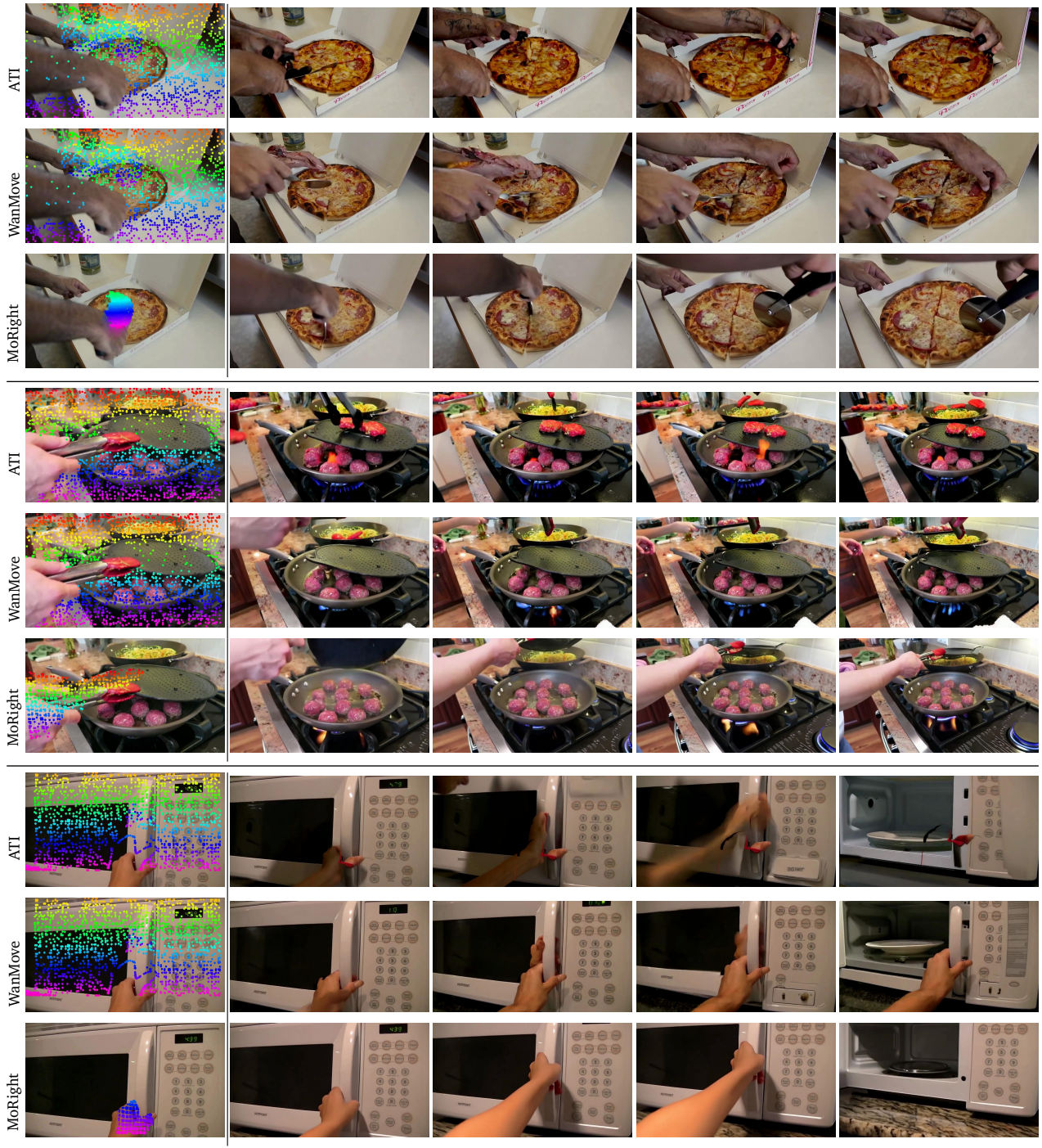

\centering
\setlength{\tabcolsep}{1pt}
\resizebox{\linewidth}{!}{%
\begin{tabular}{@{}c c|cccc@{}}
\raisebox{1.5em}{\rotatebox[origin=c]{90}{\scriptsize ATI}} &
\visrow{comparison}{ATI/5ba877dff47d6d57a7c36df616851a032e4e16aa2bbf784ffc25379d86e02f14} \\[1pt]
\raisebox{1.5em}{\rotatebox[origin=c]{90}{\scriptsize WanMove}} &
\visrow{comparison}{WanMove/5ba877dff47d6d57a7c36df616851a032e4e16aa2bbf784ffc25379d86e02f14} \\[1pt]
\raisebox{1.5em}{\rotatebox[origin=c]{90}{\scriptsize \ours}} &
\visrow{comparison}{ours/5ba877dff47d6d57a7c36df616851a032e4e16aa2bbf784ffc25379d86e02f14}{} \\

\midrule

\raisebox{1.5em}{\rotatebox[origin=c]{90}{\scriptsize ATI}} &
\visrow{comparison}{ATI/6e68609f-b587-40cc-baff-37e66947fdeb} \\[1pt]
\raisebox{1.5em}{\rotatebox[origin=c]{90}{\scriptsize WanMove}} &
\visrow{comparison}{WanMove/6e68609f-b587-40cc-baff-37e66947fdeb} \\[1pt]
\raisebox{1.5em}{\rotatebox[origin=c]{90}{\scriptsize \ours}} &
\visrow{comparison}{ours/6e68609f-b587-40cc-baff-37e66947fdeb}{} \\

\midrule

\raisebox{1.5em}{\rotatebox[origin=c]{90}{\scriptsize ATI}} &
\visrow{comparison}{ATI/b98c53a1-d724-43f2-8771-cafcb9943b84} \\[1pt]
\raisebox{1.5em}{\rotatebox[origin=c]{90}{\scriptsize WanMove}} &
\visrow{comparison}{WanMove/b98c53a1-d724-43f2-8771-cafcb9943b84} \\[1pt]
\raisebox{1.5em}{\rotatebox[origin=c]{90}{\scriptsize \ours}} &
\visrow{comparison}{ours/b98c53a1-d724-43f2-8771-cafcb9943b84}{} \\

\end{tabular}%
}
\caption{\footnotesize
\textbf{Additional qualitative comparison} with ATI~\cite{wang2025ati}, WanMove~\cite{chu2025wanmove}, and \ours.
ATI and WanMove rely on privileged 3D trajectories (with depth) projected to pixel-aligned per-frame tracks and take full interaction trajectories (active and passive) as input. In contrast, \ours uses only first-frame active tracks without privileged information and infers plausible interactions.}
\label{fig:supp_qualitative_compare}
\end{figure}

\section{More Qualitative Results}
\label{sec:more_qualitative}

\subsection{Interactive Motion Generation}
We demonstrate the causal reasoning ability of our model in \cref{fig:supp_reasoning}. We generate videos and select the first-stream generation (static view) to visualize the interaction dynamics. The input motion tracks are overlaid on the generated frames, where colored tracks indicate either user actions (active motion) or passive trajectories. Our model supports both forward and inverse reasoning. In forward reasoning (first two samples), the model predicts plausible scene consequences given the specified active motion. In inverse reasoning (last sample), the model infers feasible driving actions that could lead to the observed passive outcomes. These examples highlight the model's ability to reason about causal interactions between actions and objects.

\subsection{Disentangled Camera-Object Control}

We present additional disentangled controllable generation results in \cref{fig:supp_control}. Object motion trajectories are overlaid on the input image. We demonstrate three different object motions and three different camera motions (orbit-left, zoom-in, and zoom-out), resulting in 9 generated videos in 3 group. Each group shares the same object motion while varying the camera viewpoint, highlighting the model's ability to maintain consistent object dynamics under different camera controls. Minor variations under the same object motion but different camera motions arise from the stochastic nature of interaction generation.

\subsection{Qualitative Comparison}

We provide additional qualitative comparisons with ATI~\cite{wang2025ati} and WanMove~\cite{chu2025wanmove} in \cref{fig:supp_qualitative_compare}. Both baselines rely on privileged 3D trajectories (with depth) projected to pixel-aligned per-frame tracks and take full interaction trajectories (active and passive) as input. In contrast, our method only requires 2D motion trajectories on the first frame, while camera motion is introduced in the second stream of our dual-stream architecture. Despite using weaker inputs, our approach achieves stronger controllability and produces more coherent interactions while maintaining disentangled camera–object motion.

\clearpage
{
\small
\bibliographystyle{abbrv}
\bibliography{arxiv}
}

\end{document}